\documentclass{article}

\usepackage[main, final]{neurips_2026}

\usepackage[utf8]{inputenc}
\usepackage[T1]{fontenc}
\usepackage{hyperref}
\usepackage{url}
\usepackage{booktabs}
\usepackage{amsfonts}
\usepackage{amsmath,amssymb,mathtools}
\usepackage{nicefrac}
\usepackage{microtype}
\usepackage[table]{xcolor}
\usepackage{graphicx}
\usepackage{multirow}
\usepackage{wrapfig}
\usepackage{subcaption}
\usepackage{algorithm}
\usepackage{algpseudocode}
\usepackage{cleveref}
\usepackage{wrapfig}
\usepackage{caption}
\usepackage[T1]{fontenc}

\usepackage{mathtools}
\usepackage{amsmath, amssymb, graphicx, bm}
\usepackage[dvipsnames]{xcolor}
\usepackage{multirow}
\usepackage[table]{xcolor}
\usepackage{colortbl}
\usepackage{tcolorbox}
\usepackage{listings}
\usepackage{animate}
\usepackage{amsthm}
\newtheorem{proposition}{Proposition}

\newtheorem{manualpropositioninner}{Proposition}
\newenvironment{manualproposition}[1]{%
  \renewcommand\themanualpropositioninner{#1}%
  \manualpropositioninner
}{\endmanualpropositioninner}

\tcbuselibrary{listings, skins}
\newtcblisting{promptbox}[1]{
    colback=gray!5, 
    colframe=gray!80, 
    fonttitle=\bfseries\sffamily,
    coltitle=white,
    title=#1,
    enhanced,
    attach boxed title to top left={yshift=-2mm, xshift=5mm},
    boxed title style={colback=gray!80},
    listing only,
    listing options={
        language=Python,
        basicstyle=\ttfamily\small,
        breaklines=true,
        columns=fullflexible,
        commentstyle=\color{green!50!black},
        keywordstyle=\color{blue},
        stringstyle=\color{orange},
        showstringspaces=false,
        morekeywords={role, content, system, user, assistant},
        deletekeywords={in}
    }
}
\usepackage{paralist}
\usepackage{wrapfig}
\usepackage[rflt]{floatflt}

\definecolor{mygreen}{RGB}{0,180,0}
\definecolor{mcfg_red}{HTML}{E63946}
\definecolor{ig_orange}{HTML}{FFB800}
\definecolor{scfg_green}{HTML}{2A9D8F}
\definecolor{cads_blue}{HTML}{1D3557}
\colorlet{lightgreen}{mygreen!15}

\def\eqref#1{Eq.~(\ref{#1})}
\def\bx{\boldsymbol{x}}
\def\bz{\boldsymbol{z}}

\def\bn{\boldsymbol{n}}
\def\bc{\boldsymbol{c}}

\RequirePackage{xspace}
\makeatletter
\DeclareRobustCommand\onedot{\futurelet\@let@token\@onedot}
\def\@onedot{\ifx\@let@token.\else.\null\fi\xspace}

\def\eg{\emph{e.g}\onedot} 

\def\ie{\emph{i.e}\onedot}

\def\etal{\emph{et al}\onedot}
\makeatother

\providecommand{\eg}{\emph{e.g.}}
\providecommand{\ie}{\emph{i.e.}}
\providecommand{\etal}{\emph{et al.}}

\definecolor{lightgreen}{RGB}{220,245,220}

\title{MotionCFG: Boosting Motion Dynamics\\via Semantic Motion Sharpening}

\author{%
  Byungjun Kim \\
  KAIST \\
  Daejeon 34141, Republic of Korea \\
  \texttt{bjkim@kaist.ac.kr} \\
  \AND
  Soobin Um$^{\dagger}$ \\
  DGIST \\
  Daegu 42988, Republic of Korea \\
  \texttt{soobin.um@dgist.ac.kr} \\
  \And
  Jong Chul Ye$^{\dagger}$ \\
  KAIST \\
  Daejeon 34141, Republic of Korea \\
  \texttt{jong.ye@kaist.ac.kr} \\
}

\begin{document}

\maketitle

\begingroup
  \renewcommand{\thefootnote}{\fnsymbol{footnote}}
  \footnotetext[2]{Corresponding authors.}
\endgroup

\begin{abstract}
Despite recent advancements in Text-to-Video (T2V) synthesis, generating high-fidelity and dynamic motion remains a significant challenge. Existing methods primarily rely on Classifier-Free Guidance (CFG), but standard CFG treats all semantic dimensions uniformly and provides no mechanism to selectively resolve ambiguity along the motion axis. 
To address this, we propose MotionCFG, a training-free approach that introduces selective guidance along the motion subspace of the condition embedding. Specifically, we construct a motion-perturbed negative condition by injecting Gaussian noise into motion-related embeddings, and steer generation away from it to selectively amplify motion intent. We show theoretically that this procedure implicitly approximates \textit{Semantic Laplacian Sharpening}---a second-order correction that suppresses motion-ambiguous regions of the score landscape and amplifies well-resolved dynamic peaks.
Combined with a piecewise guidance schedule that confines intervention to the early denoising steps, MotionCFG consistently improves motion dynamics across state-of-the-art T2V frameworks with negligible overhead. We further demonstrate that this Laplacian sharpening principle generalizes beyond motion, effectively steering complex, non-linear concepts such as precise object numerosity that are typically difficult to modulate via standard text-based guidance. Our project page is available on \textcolor{RubineRed}{\fontfamily{lmtt}\selectfont{https://motion-cfg.vercel.app/}}.
\end{abstract}

\begin{figure}[t]
    \centering
    \includegraphics[width=1.0\linewidth]{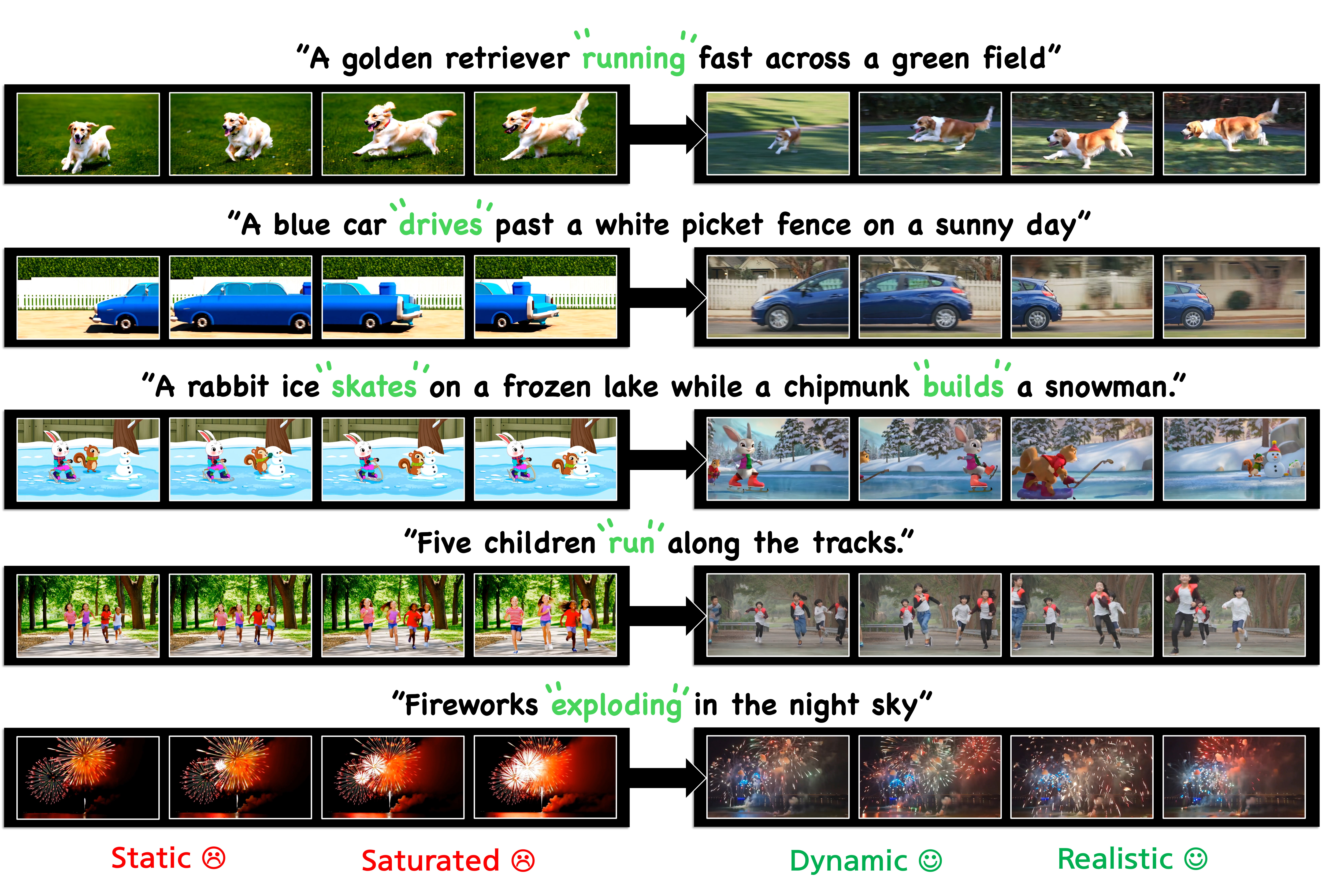}
    \vspace{-0.6cm}
    \caption{\textbf{Showcase of MotionCFG.} While Standard CFG (left) produces over-saturated colors and static outputs that often fail to reflect the intended actions, MotionCFG (right) resolves motion ambiguity by selectively sharpening motion-related embeddings, thereby yielding realistic and physically dynamic videos faithful to the prompt. Motion words are highlighted in \textcolor{green}{green}.}
    \label{fig:showcase}
    \vspace{-0.25cm}
\end{figure}

\section{Introduction}
\label{sec:intro}

Text-to-Video (T2V) generation has seen remarkable progress with the advent of large-scale diffusion-based generative frameworks~\cite{brooks2024video,chen2023videocrafter1,yang2024cogvideox,wan2025wan}, enabling the creation of realistic videos that faithfully reflect user-intended text descriptions. A key contributor to this success is Classifier-Free Guidance (CFG)~\cite{ho2022classifier}, which strengthens text-conditioning signals by extrapolating the score function away from an unconditional null embedding.

Despite this progress, recent studies consistently report that T2V models struggle to produce videos with strong dynamic motion, often yielding near-static outputs even for explicitly motion-rich prompts (\eg, ``a cat jumping over a fence'')~\cite{ruan2024enhancing,flowmo2025}. We attribute this to the indiscriminate nature of standard CFG: since it extrapolates the entire condition vector at once, semantic axes that are under-represented in the text encoder, such as motion~\cite{ruan2024enhancing,Momeni_2023_ICCV}, can remain under-resolved when competing with more dominant dimensions such as object identity or scene layout. Incorporating explicit negative prompts (\eg, ``static'') is a natural attempt, but often offers marginal gains; see Table~\ref{tab:ablation_perturbation} for instance.

In this work, we propose \textbf{MotionCFG}, a simple-yet-powerful framework that addresses motion ambiguity in a zero-shot, training-free fashion. While standard CFG guides generation uniformly away from an unconditional null embedding, MotionCFG introduces a \textit{motion-targeted} guidance mechanism that operates selectively on the motion subspace of the condition. During inference, we identify motion-related tokens (\eg, action verbs) from the input prompt, either manually or via a lightweight large language model (LLM), and inject Gaussian noise only into the embeddings of these motion tokens. The resulting motion-perturbed condition serves as the negative anchor, and we steer generation \textit{away} from this perturbed neighborhood to selectively sharpen motion semantics without affecting other attributes. To balance motion dynamics with visual fidelity, we further introduce a Piecewise Guidance Scheduler that restricts our intervention to the early sampling steps, where the global motion layout is determined.

Beyond its simplicity, MotionCFG admits an interpretable theoretical viewpoint. We show that the procedure implicitly approximates the Laplacian of the score (or velocity~\cite{lipman2022flow}) field restricted to the motion subspace of the condition. We refer to this principle as \textit{Semantic Laplacian Sharpening}: a correction that suppresses motion-ambiguous semantic regions and amplifies well-resolved dynamic peaks, extending prior data-space Laplacian correction~\cite{anumasa2025laplacian} to the condition space. This perspective also predicts that the same mechanism generalizes to other ambiguous semantic dimensions beyond motion, which we empirically verify on object numerosity (See Section~\ref{sec:counting}).

We validate MotionCFG on state-of-the-art T2V models including Wan2.1~\cite{wan2025wan} (1.3B and 14B) and CogVideoX~\cite{yang2024cogvideox}, demonstrating consistent improvements in motion dynamics with marginal compromise in text-video alignment, all at negligible computational overhead over standard CFG. Furthermore, consistent with our theoretical framework, we show that the same Semantic Laplacian Sharpening principle extends beyond motion to other notoriously difficult-to-control attributes such as precise object numerosity~\cite{binyamin2025make}, establishing MotionCFG as a general approach for resolving fine-grained semantic ambiguity in text-conditioned generation.

\section{Related Work}
\label{sec:related_work}

\noindent\textbf{Text-to-Video Generation.}
Text-to-Video (T2V) generation has evolved rapidly, transitioning from 3D-UNet based diffusion models~\cite{ho2022video, singer2022make} to Flow-Matching (FM) frameworks~\cite{lipman2022flow} utilizing Diffusion Transformers (DiT). State-of-the-art open-source models like CogVideoX~\cite{yang2024cogvideox} and WAN~\cite{wan2025wan} leverage large-scale datasets and 3D causal VAEs to achieve high spatiotemporal consistency. Despite their impressive visual quality, these models heavily rely on standard cross-attention mechanisms for text conditioning, which often fail to parse fine-grained physical dynamics, leading to motion ambiguity in complex motion generation.

\noindent\textbf{Motion Dynamics in Video Generation.}
Current text-to-video (T2V) models frequently suffer from a static bias, where generated outputs exhibit minimal temporal variation or resemble static images with subtle camera drift~\cite{ruan2024enhancing,sobolev2025state}. To address this, DEMO~\cite{ruan2024enhancing} enhances motion synthesis by decomposing text conditioning into content and motion components, but requires specialized training with text-motion and video-motion supervision. Recent training-free approaches attempt to improve motion dynamics during inference. State Guidance~\cite{sobolev2025state} introduces frame-wise scheduled guidance over state-transition prompts to encourage temporal scene progression, while VideoGuide~\cite{lee2025videoguide} enhances temporal consistency through teacher-guided denoising trajectory interpolation. In contrast, our MotionCFG directly perturbs motion-related semantic embeddings to construct a motion-aware negative branch for classifier-free guidance. MotionCFG operates entirely at inference time without requiring architectural modifications, additional optimization with memory overhead, or auxiliary teacher models, while remaining applicable to general T2V generation settings.

\noindent\textbf{Guidance Mechanisms in Diffusion Models.} Classifier-Free Guidance (CFG) is the de facto standard for aligning diffusion outputs with text prompts~\cite{ho2022classifier}. While effective for static attributes, it struggles with motion ambiguity since extrapolating away from an empty string ($\varnothing$) does not inherently resolve the specific dynamic intent. Recent works have explored explicit negative prompts, gradient-based guidance, or attention-map manipulations. Orthogonal to these approaches, Laplacian Score Sharpening~\cite{anumasa2025laplacian} shows that subtracting the Laplacian of the score can sharpen overly smooth score landscapes and suppress ambiguous inter-mode regions. Motivated by this perspective, we extend score sharpening from the data space to the semantic condition space: MotionCFG constructs its negative branch by perturbing motion-related text embeddings, yielding an efficient inference-time approximation to a Laplacian correction along motion-relevant semantic directions.

\section{Method}
\label{sec:method}

\subsection{Preliminaries}
\label{sec:prelim}

\noindent \textbf{Text-to-Video Diffusion Models.}
Modern diffusion-based text-to-video (T2V) models~\cite{brooks2024video,chen2023videocrafter1,yang2024cogvideox,wan2025wan} operate in the latent space of a 3D variational autoencoder (VAE)~\cite{kingma2013auto}, which consists of an encoder $E(\cdot)$ and a decoder $D(\cdot)$. A video $\bx$ is first compressed into a low-dimensional latent representation $\bz_0 = E(\bx)$. The diffusion forward process gradually corrupts $\bz_0$ by adding Gaussian noise according to a variance schedule, producing a noisy latent $\bz_t$ at timestep $t$~\cite{ho2020denoising}. A neural network is then trained to reverse this corruption by learning to predict the added noise $\epsilon_\theta(\bz_t, \bc)$~\cite{ho2020denoising}, where $\bc$ denotes the text condition embedded by a pretrained text encoder (\eg, T5~\cite{raffel2020exploring}). Generation proceeds by iteratively denoising from pure noise $\bz_T \sim \mathcal{N}(\mathbf{0}, \mathbf{I})$ back to a clean latent $\bz_0$ using the learned noise predictor, which is then decoded to pixel space via $\hat{\bx} = D(\bz_0)$.

\noindent \textbf{Classifier-Free Guidance.}
Classifier-Free Guidance (CFG)~\cite{ho2022classifier} improves text-conditional generation by extrapolating between the unconditional and conditional noise predictions:
\begin{equation}
    \tilde{\epsilon}_\theta(\bz_t, \bc, \varnothing) = \epsilon_\theta(\bz_t, \varnothing) + \omega_{\text{std}} \left( \epsilon_\theta(\bz_t, \bc) - \epsilon_\theta(\bz_t, \varnothing) \right),
    \label{eq:cfg}
\end{equation}
where $\varnothing$ denotes the null (unconditional) embedding and $\omega_{\text{std}} > 1$ is the guidance scale. By extrapolating the noise prediction away from the unconditional estimate, CFG sharpens the conditional distribution and strengthens text alignment. However, this extrapolation operates on the entire condition vector $\bc$ uniformly. Since the condition embedding encodes multiple semantic axes simultaneously (\eg, object identity, scene layout), fine-grained axes such as motion can remain under-resolved when the extrapolation is dominated by other, more salient dimensions. This motivates a guidance mechanism that can selectively target the motion subspace of the condition.

A natural first attempt is \emph{negative prompting}~\cite{desai2024improving}: replacing the null embedding in~\eqref{eq:cfg} with a text condition that explicitly describes static or low-motion scenes (\eg, ``static, still picture''), thereby steering the generation away from stillness. In practice, however, this coarse textual intervention yields only marginal gains over standard null-prompted CFG (see Table~\ref{tab:ablation_perturbation}), as a single hand-crafted sentence cannot capture the fine-grained motion semantics distributed across individual tokens.

\begin{figure}[t]
    \centering
    \includegraphics[width=1.0\linewidth]{figures/concept.jpg}
    \caption{
        \textbf{Overview of the MotionCFG pipeline.} 
        \textbf{Step 1:} Motion-related tokens in the prompt are identified via an LLM. 
        \textbf{Step 2:} Gaussian noise ($\delta_t$) is injected exclusively into these motion text embeddings to generate a perturbed condition ($c_{pert,t}$). 
        \textbf{Step 3:} A piecewise guidance schedule applies MotionCFG during the early sampling steps to establish robust motion trajectories, before reverting to standard CFG to refine spatial details.
    }
    \label{fig:concept_fig}
    \vspace{-0.2cm}
\end{figure}

\subsection{MotionCFG}
\label{sec:motioncfg}

We introduce MotionCFG, a training-free sampler that addresses the under-resolution of the motion axis by incorporating a motion-perturbed negative anchor. Specifically, given an input text prompt, let $\bc \in \mathbb{R}^{L \times d}$ denote its embedding sequence obtained from the pretrained text encoder, where $L$ is the sequence length and $d$ is the embedding dimension. We first identify a set of motion-related tokenized indices $\mathcal{I}_{\text{motion}}$ (\eg, action verbs such as ``jumping'' or ``running''). At each denoising step $t$, we sample a fresh noise vector $\bn_t \sim \mathcal{N}(\mathbf{0}, \mathbf{I})$ and construct a perturbed embedding $\bc_{\text{pert},t}$ by injecting noise exclusively into the text embeddings at the identified indices:
\begin{equation}
    \bc_{\text{pert},t}[i] = 
    \begin{cases} 
        \bc[i] + \sigma_{\bc}\,\bn_t, & \text{if } i \in \mathcal{I}_{\text{motion}} \\
        \bc[i], & \text{otherwise}
    \end{cases}
    \label{eq:perturbation}
\end{equation}
where $\sigma_{\bc} > 0$ controls the perturbation magnitude. 
The guided noise prediction is then computed by steering generation away from this perturbed neighborhood:
\begin{equation}
    \tilde{\epsilon}_\theta(\bz_t, \bc, \bc_{\text{pert},t}) = \epsilon_\theta(\bz_t, \bc_{\text{pert},t}) + \omega \left( \epsilon_\theta(\bz_t, \bc) - \epsilon_\theta(\bz_t, \bc_{\text{pert},t}) \right),
    \label{eq:motioncfg}
\end{equation}
where $\bc$ and $\bc_{\text{pert},t}$ denote the conditions derived from the original and perturbed embeddings, respectively. 
Note that the anchor point is different from standard CFG in \eqref{eq:cfg}.
Intuitively, while standard CFG extrapolates between the \emph{unconditional} and conditional noise predictions to strengthen overall text alignment, MotionCFG steers away from a \emph{motion-degraded} condition, thereby selectively sharpening the motion semantics without affecting the remaining attributes.

\noindent \textbf{Why perturb motion embeddings?}
MotionCFG intervenes on the final contextualized T5 embeddings received by the denoiser. The selected positions provide entry points for the requested action, while motion arising from scene context, object interactions, camera behavior, and temporal dynamics remains modeled by the pretrained video generator. To examine whether these positions retain traces of the selected action tokens after contextualization, let $\mathbf{e}_i$ denote the input embedding at motion-token position $i$ and $\mathbf{c}_j$ the final contextual embedding at position $j$. We measure each row's maximum cosine similarity to the selected input embeddings
$s_j = \max_{i \in \mathcal{I}_{\text{motion}}}
\cos(\mathbf{e}_i,\mathbf{c}_j).
\label{eq:motion_token_similarity}
$

\begin{wraptable}[13]{r}{0.55\textwidth}
\vspace{-0.4cm}
\centering
\caption{\textbf{Motion-token localization.}
We compare final contextual T5 rows with the selected input-token embeddings.
$\operatorname{TopK}(\mathbf{s})$ denotes the indices of the $K$ highest-scoring rows, where $K=|\mathcal{I}_{\text{motion}}|$ and $\mathbf{s}=(s_1,\ldots,s_L)$.}
\label{tab:motion_token_localization}
\footnotesize
\setlength{\tabcolsep}{3pt}
\resizebox{0.55\textwidth}{!}{%
\begin{tabular}{lcc}
\toprule
\textbf{Measurement} & \textbf{Definition} & \textbf{Result} \\
\midrule
Motion-row similarity
& $\frac{1}{K}\sum_{j\in\mathcal{I}_{\text{motion}}}s_j$
& 0.108 \\
Other-row similarity
& $\frac{1}{L-K}\sum_{j\notin\mathcal{I}_{\text{motion}}}s_j$
& 0.014 \\
Top-1 localization
& $\mathbf{1}\!\left[
\operatorname*{arg\,max}_{j}s_j\in\mathcal{I}_{\text{motion}}
\right]$
& $100\%$ \\
Top-$K$ recall
& $\frac{|\operatorname{TopK}(\mathbf{s})
\cap\mathcal{I}_{\text{motion}}|}{K}$
& $88.9\%$ \\
\bottomrule
\end{tabular}%
}
\vspace{-1em}
\end{wraptable}

Table~\ref{tab:motion_token_localization} shows that the selected rows retain stronger traces of the original motion-token inputs despite contextual mixing. The highest-scoring row occurs at a selected position for every evaluated prompt, and most selected positions are recovered among the top-$K$ rows. These measurements support the selected contextual rows as practical anchors for intervening on action-related conditioning, without requiring motion to be fully localized or disentangled at those positions. Appendix~\ref{sec:apdx_ablations} provides a controlled comparison of perturbing initial T5 input embeddings versus final contextual embeddings.

\begin{figure}[t]
\centering

\begin{minipage}[t]{0.53\linewidth}
\vspace{0pt}
\centering
\includegraphics[width=\linewidth]{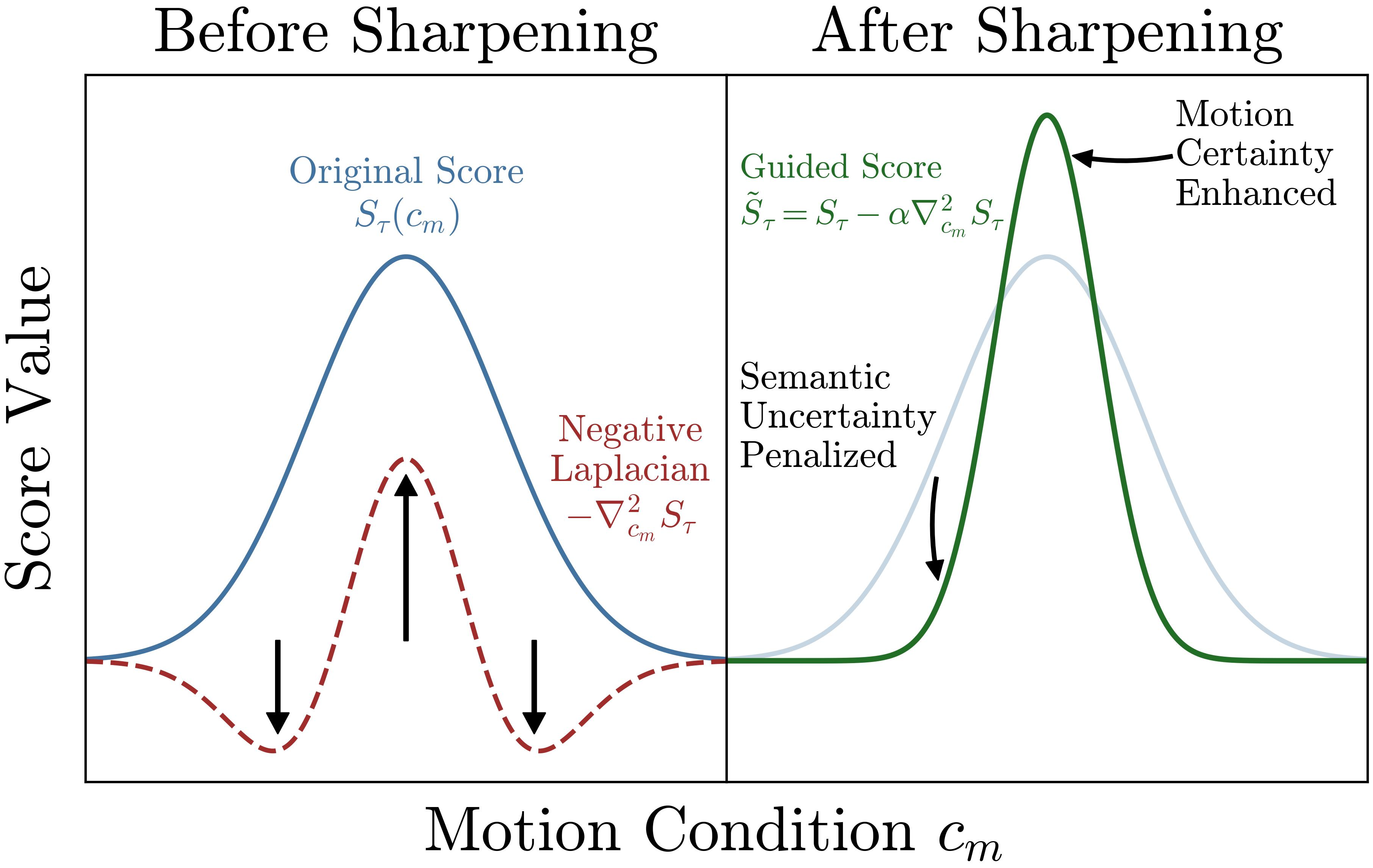}
\captionof{figure}{
\textbf{Semantic Laplacian Sharpening.}
Laplacian sharpening suppresses motion-ambiguous score regions and sharpens confident motion modes.
}
\label{fig:sharpening}
\end{minipage}
\hfill
\begin{minipage}[t]{0.45\linewidth}
\vspace{-0.3cm}
\begin{algorithm}[H]
\caption{MotionCFG Sampling}
\label{alg:motioncfg}
\scriptsize
\begin{algorithmic}[1]
\State \textbf{Input:} $\mathcal{P}, \epsilon_\theta, \mathcal{D}, T, \tau, \omega, \omega_{\text{std}}$
\State $\mathcal{I}_{\text{motion}} \leftarrow \text{MotionExtract}(\mathcal{P})$
\State $\bc \leftarrow \text{Encode}(\mathcal{P})$
\State $\bc_{\text{neg}} \leftarrow \text{Encode}(\mathcal{P}_{\text{neg}})$
\State $\bz_T \sim \mathcal{N}(\mathbf{0}, \mathbf{I})$
\For{$t = T \dots 1$}
    \If{$t > (1-\tau)T$}
        \State $\bn_t \sim \mathcal{N}(\mathbf{0}, \mathbf{I})$
        \State $\bc_{\text{pert},t} \leftarrow \text{Perturb}(\bc,\mathcal{I}_{\text{motion}},\sigma_{\bc}\bn_t)$
        \State $\tilde{\epsilon} \leftarrow \epsilon_\theta(\bz_t,\bc)$
        \Statex $\quad\quad\quad\quad\quad\quad +\,\omega\big(\epsilon_\theta(\bz_t,\bc)-\epsilon_\theta(\bz_t,\bc_{\text{pert},t})\big)$
    \Else
        \State $\tilde{\epsilon} \leftarrow \epsilon_\theta(\bz_t,\varnothing)$
        \Statex $\quad\quad\quad\quad\quad\quad +\,\omega_{\text{std}}\big(\epsilon_\theta(\bz_t,\bc)-\epsilon_\theta(\bz_t,\varnothing)\big)$
    \EndIf
    \State $\bz_{t-1} \leftarrow \text{SchedulerStep}(\bz_t,\tilde{\epsilon})$
\EndFor
\State \textbf{Return} $\hat{\bx} \leftarrow \mathcal{D}(\bz_0)$
\end{algorithmic}
\end{algorithm}
\end{minipage}

\vspace{-0.3em}
\end{figure}

\noindent \textbf{Piecewise Guidance Scheduling.}
In video diffusion models, the global motion layout is established during the earliest denoising steps, while later steps refine high-frequency spatial details and textures~\cite{baherwani2025characterizing,jang2026frame}. Applying~\eqref{eq:motioncfg} throughout the entire sampling process introduces unnecessary stochastic variance in the late stage, often resulting in visual artifacts. We therefore introduce a piecewise schedule controlled by a ratio $\tau \in [0, 1]$: MotionCFG is applied during the first $\tau$ fraction of the total steps to establish the motion trajectory, after which we revert to standard CFG (\eqref{eq:cfg}) for spatial refinement.
 To jointly improve text alignment during the motion-encouraging stage, we anchor the baseline on $\bc$ rather than $\bc_{\text{pert},t}$:
\begin{equation}
    \tilde{\epsilon}_\theta^*(\bz_t, \bc, \bc_{\text{pert},t}) = \epsilon_\theta(\bz_t, \bc) + \omega \left( \epsilon_\theta(\bz_t, \bc) - \epsilon_\theta(\bz_t, \bc_{\text{pert},t}) \right).
    \label{eq:motioncfg_pw}
\end{equation}
This two-stage approach with the $\bc$-anchored formulation consistently outperforms the full-schedule variant ($\tau = 1.0$); see Figure~\ref{fig:tradeoff} for details. The full procedure is summarized in Algorithm~\ref{alg:motioncfg}, and an overview of the pipeline is illustrated in Figure~\ref{fig:concept_fig}.

\noindent \textbf{Motion Embeddings Identification.}
Identifying $\mathcal{I}_{\text{motion}}$ requires determining which parts of the text embeddings in the prompt carry motion semantics. Users can manually specify target verbs for best precision, but this becomes tedious when processing large prompt sets. To sidestep this, we automate the identification by querying a lightweight large language model (\eg, Qwen2.5-7B-Instruct~\cite{qwen2025qwen25technicalreport}) via in-context learning. Specifically, we feed the prompt with its tokenized indices (\eg, \texttt{[0]A [1]cat [2]jumps}) and instruct the model to return the indices of words describing physical actions or dynamic changes. Importantly, the computational overhead introduced by this automated extraction is marginal compared to the overall video diffusion process. Detailed complexity analysis, along with the instruction template, is provided in the supplementary.

\subsection{Connection to Semantic Laplacian Sharpening}
\label{sec:theory}

Here we provide a theoretical insight into why MotionCFG is effective. In particular, we show that MotionCFG can be interpreted as an efficient, single-sample approximation of \emph{Semantic Laplacian Sharpening}, a second-order correction that sharpens the score landscape along the motion dimensions of the condition space. For convenience, here we work with the score notation $\mathbf{S}_\theta \coloneqq -\boldsymbol{\epsilon}_\theta / \sigma_t$~\cite{song2020score, dhariwal2021diffusion}, which is equivalent to the noise prediction $\boldsymbol{\epsilon}_\theta$ up to a timestep-dependent scalar determined by the noise schedule $\sigma_t$. Under this notation, the MotionCFG update in~\eqref{eq:motioncfg_pw} becomes
\begin{equation}
    \tilde{\mathbf{S}}_\theta^*(\bz_t, \bc, \bc_{\text{pert},t}) = \mathbf{S}_\theta(\bz_t, \bc) + \omega \left( \mathbf{S}_\theta(\bz_t, \bc) - \mathbf{S}_\theta(\bz_t, \bc_{\text{pert},t}) \right).
    \label{eq:motioncfg_pw_score}
\end{equation}

To establish the connection to Laplacian sharpening, we first consider regularizing the score function by subtracting its Laplacian with respect to the motion condition:
\begin{equation}
    \mathbf{S}^*_\theta(\bz_t, \bc) = \mathbf{S}_\theta(\bz_t, \bc) - \gamma\,\nabla_{\bc_m}^2\,\mathbf{S}_\theta(\bz_t, \bc),
    \label{eq:semantic_laplacian}
\end{equation}
where $\bc_m \in \mathbb{R}^{|\mathcal{I}_{\text{motion}}| \times d}$ denotes the rows of $\bc$ indexed by $\mathcal{I}_{\text{motion}}$, and $\gamma > 0$ controls the regularization strength. Intuitively, the Laplacian $\nabla_{\bc_m}^2 \mathbf{S}_\theta$ measures how sensitive the score is to variations in the motion condition: it is near-zero in flat, ambiguous regions where multiple motion interpretations coexist, and strongly negative at sharp peaks where the motion intent is well-resolved. By subtracting this term, \eqref{eq:semantic_laplacian} penalizes the former and amplifies the latter, effectively sharpening the score landscape along the motion axis. This extends the Laplacian Score Sharpening of Chandran~\etal~\cite{anumasa2025laplacian}, who apply an analogous correction in the \emph{data space} to reduce visual hallucinations, to the \emph{condition space}, hence the name \emph{Semantic} Laplacian Sharpening. See Figure~\ref{fig:sharpening} for illustration of the concept.

The problem is that computing $\nabla_{\bc_m}^2\,\mathbf{S}_\theta$ exactly requires evaluating the trace of the Hessian via second-order backpropagation through the denoising network, which is prohibitively expensive for large-scale video models. Remarkably, MotionCFG bypasses this entirely, offering an efficient proxy for~\eqref{eq:semantic_laplacian} as formalized below.
\begin{proposition}
\label{prop:laplacian}
Let $\bn_t \sim \mathcal{N}(\mathbf{0}, \mathbf{I})$ be the noise realization in~\eqref{eq:perturbation}. Assuming $\mathbf{S}_\theta(\bz_t, \cdot)$ is three-times differentiable with respect to the condition, the expected guided score in~\eqref{eq:motioncfg_pw_score} satisfies
\begin{equation}
    \mathbb{E}_{\bn_t}\left[
    \tilde{\mathbf{S}}_\theta^*(\bz_t, \bc, \bc_{\text{pert},t})
    \right] \;=\; \mathbf{S}_\theta(\bz_t, \bc) \;-\; \frac{\omega\,\sigma_{\bc}^2}{2}\,\nabla_{\bc_m}^2\,\mathbf{S}_\theta(\bz_t, \bc) \;+\; O(\sigma_{\bc}^3),
    \label{eq:laplacian_result}
\end{equation}
where $\nabla_{\bc_m}^2$ is the Laplacian with respect to $\bc_m$.
\end{proposition}
See Section~\ref{app:math_derivation} for the proof. Notice that~\eqref{eq:laplacian_result} recovers the Semantic Laplacian Sharpening target in~\eqref{eq:semantic_laplacian} with $\gamma = \omega\sigma_{\bc}^2/2$, up to third-order residuals. This means that the MotionCFG update in~\eqref{eq:motioncfg_pw_score} is interpretable as an unbiased estimator of the Laplacian-regularized score, requiring only one extra forward pass (\ie, the same cost as standard CFG).

While Proposition~\ref{prop:laplacian} holds in expectation under idealized assumptions, the practical MotionCFG update involves several approximations, including our use of single-sample Monte Carlo for $\bn_t$ and the residual term $O(\sigma_{\bc}^3)$. To assess whether the Laplacian sharpening interpretation remains meaningful in practice, we conduct an empirical analysis on synthetic toy data, finding that the qualitative sharpening behavior predicted by the framework is preserved across our practical design choices. See Section~\ref{sec:toy_experiment} for details.

\section{Experiments}
\label{sec:experiments}

\begin{figure}[t]
    \centering
    \includegraphics[width=1.0\linewidth]{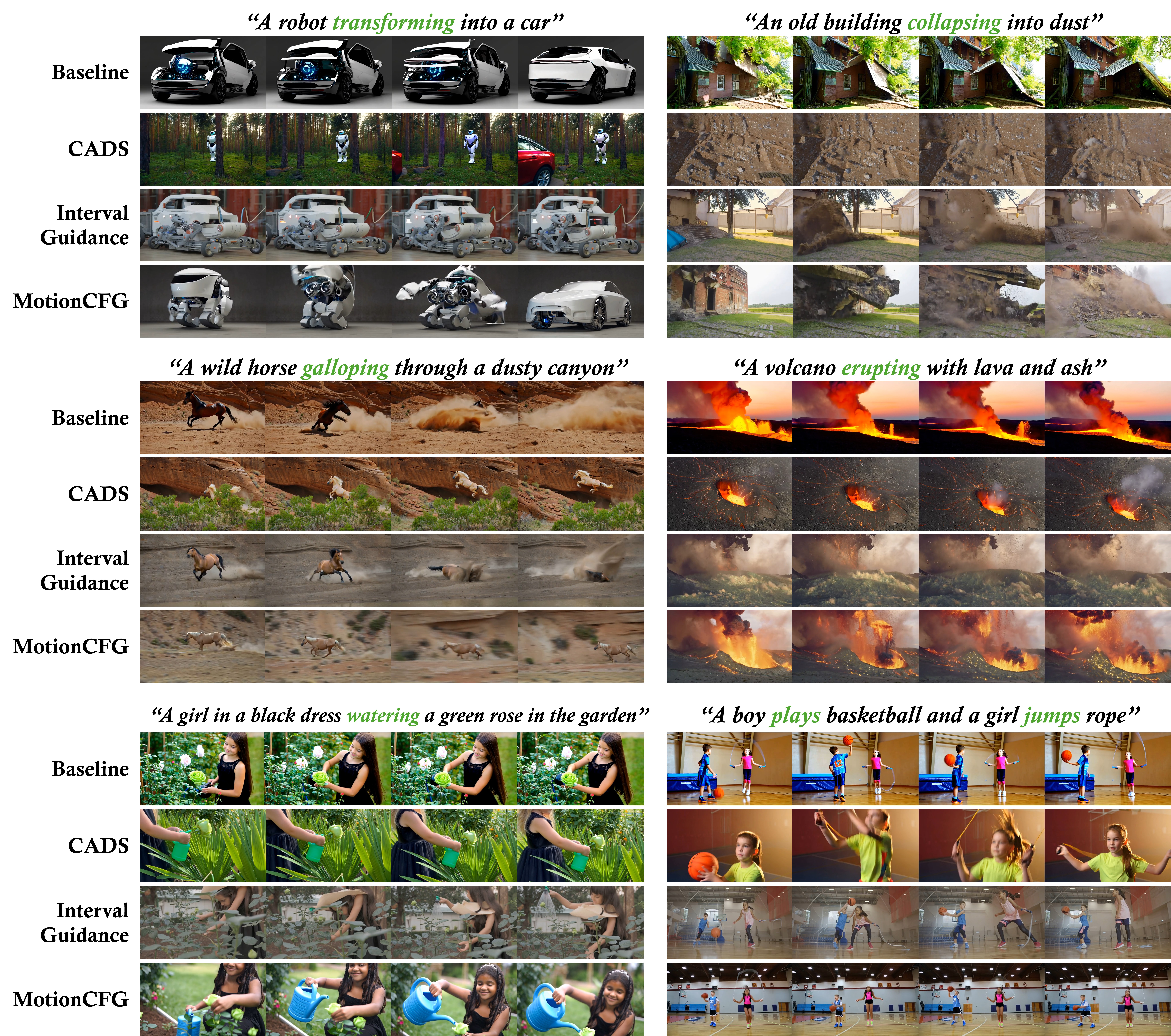}

    \caption{\textbf{Qualitative comparison on Wan2.1.} Each row shows uniformly sampled frames from a generated video. ``Baseline'' denotes negative-prompted CFG. While existing approaches produce near-static outputs or visual artifacts, MotionCFG generates physically plausible dynamics faithful to the highlighted motion tokens (\textcolor{green}{green}). }
    \label{fig:qualitative}
    \vspace{-0.3cm}
\end{figure}

\noindent \textbf{Models and Datasets.} 
To verify the robustness of our approach across various model scales, we conduct experiments on several state-of-the-art T2V frameworks, including Wan2.1-T2V (1.3B and 14B) and CogVideoX-5B. For quantitative and qualitative evaluation, we utilize the T2V-Compbench~\cite{sun2024t2v} dataset. However, as T2V-Compbench does not explicitly prioritize or categorize high-magnitude motion dynamics, we curate an additional set of motion-intensive prompts using Gemini to specifically assess dynamic fidelity. A comprehensive description of these curated prompts and their categories is provided in the supplementary.

\noindent \textbf{Baselines.} 
We compare MotionCFG against several training-free sampling strategies. 
\textit{Baseline} denotes \textit{Negative Prompting}, which replaces the null embedding in standard CFG with a text condition describing static scenes (\eg, ``static, still picture''); details on prompt selection are provided in the supplementary. 
\textit{CADS}~\cite{sadat2024cads} and \textit{Interval Guidance}~\cite{kynkaanniemi2024applying} are state-of-the-art diversity-enhancing samplers originally proposed for image generation. 
We additionally compare against \textit{State Guidance (SG)}~\cite{sobolev2025state}, which applies frame-wise scheduled guidance over state-transition prompts, and \textit{VideoGuide}~\cite{lee2025videoguide}, which improves temporal consistency via teacher-guided denoising trajectory interpolation during sampling. 
We further include \textit{FlowMo}~\cite{flowmo2025}, a concurrent training-free motion-enhancing approach that performs step-wise gradient-based optimization in the latent space to reduce temporal variance. Unlike MotionCFG, FlowMo requires per-step gradient computation, introducing substantial computational overhead; see Appendix~\ref{apdx:computation} for detailed analysis.

\noindent \textbf{Evaluation Metrics.}
To comprehensively evaluate the generated videos, we employ a diverse set of metrics focusing on two main aspects: \emph{visual quality with semantic alignment}, and \emph{motion dynamics}. 
For visual quality and semantic alignment, we evaluate videos at both the frame and video levels. For frame-wise evaluation, we extract individual frames from the generated video, independently calculate \textbf{CLIPScore}~\cite{hessel2021clipscore} and \textbf{PickScore}~\cite{Kirstain2023PickaPicAO}, and report the averaged scores for each video. To assess the global spatio-temporal alignment between the text prompt and the entire video sequence, we employ \textbf{X-CLIP}~\cite{ma2022x}, a dedicated video-language understanding metric.

To quantitatively evaluate the magnitude, complexity, and temporal dynamics of the generated motions, we adopt the evaluation protocol proposed in the \textbf{DEVIL} benchmark~\cite{liao2024evaluation}. Specifically, we report \textbf{FLOW} (average pixel-level motion) and semantic variations using \textbf{Info DINO} and \textbf{DINO Segm}. We also report the unified \textbf{DEVIL Score}. Notably, this score is derived via regression after measuring a broader, comprehensive set of multiple dynamic indicators to provide a holistic assessment. Additionally, we employ \textbf{SSIM} to measure structural variation across adjacent frames, where a lower score indicates higher dynamic changes.

\begin{table}[t]
    \centering
    \caption{\textbf{Quantitative comparison on Gemini-generated prompts}. ``Baseline'' denotes negative-prompted CFG. MotionCFG (Ours) consistently demonstrates superior performance in temporal dynamics across various base models. Upward arrows ($\uparrow$) indicate higher is better, while downward arrows ($\downarrow$) indicate lower is better.}
    \label{tab:gemini_results}

    \resizebox{\textwidth}{!}{%
    \begin{tabular}{ll cccccccc}
        \toprule
        \multirow{2}{*}{\textbf{Model}} & \multirow{2}{*}{\textbf{Method}} & \multicolumn{2}{c}{\textbf{Framewise}} & \textbf{Full Video} & \multicolumn{5}{c}{\textbf{Dynamics / Temporal}} \\
        \cmidrule(lr){3-4} \cmidrule(lr){5-5} \cmidrule(lr){6-10}
        & & CLIPScore$\uparrow$ & PickScore$\uparrow$ & X-CLIP$\uparrow$ & SSIM$\downarrow$ & FLOW$\uparrow$ & Info DINO$\uparrow$ & DINO Segm.$\uparrow$ & DEVIL$\uparrow$ \\
        \midrule
        \multirow{7}{*}{Wan2.1-1.3B}
        & Baseline & 0.2535 & 0.2092 & 22.2128 & 0.7193 & 4.0962 & 0.0555 & 0.3842 & 0.3069 \\
        & CADS & 0.2496 & 0.2104 & 22.7687 & \textbf{0.6476} & 2.2151 & 0.0294 & 0.3553 & 0.1973 \\
        & IG & 0.2392 & 0.2058 & 23.0677 & 0.7116 & 3.7564 & 0.0635 & 0.4420 & 0.3147 \\
        & FlowMo & \textbf{0.2622} & \textbf{0.2120} & 22.4004 & 0.7669 & 3.5846 & 0.0447 & 0.3558 & 0.3032 \\
        & VideoGuide & 0.2554 & 0.2118 & 22.7258 & 0.6287 & 4.1330 & 0.0366 & 0.3780 & 0.2636 \\
        & SG & \textbf{0.2535} & \textbf{0.2089} & 22.2330 & 0.7359 & 3.7479 & 0.0572 & 0.3853 & 0.3071 \\
        \rowcolor{lightgreen}
        \cellcolor{white} & \textbf{Ours} & 0.2488 & 0.2089 & \textbf{23.3881} & 0.6547 & \textbf{4.7320} & \textbf{0.0650} & \textbf{0.4586} & \textbf{0.3374} \\
        \midrule
        \multirow{5}{*}{Wan2.1-14B}
        & Baseline & \textbf{0.2651} & 0.2136 & 21.7438 & 0.8398 & 2.9466 & 0.0430 & 0.2853 & 0.2324 \\
        & CADS & 0.2516 & 0.2112 & 22.1356 & 0.7735 & 2.4108 & 0.0467 & 0.3554 & 0.2410 \\
        & IG & 0.2518 & 0.2112 & \textbf{23.0622} & 0.8207 & 2.6495 & 0.0571 & 0.3931 & 0.2732 \\
        & FlowMo & 0.2626 & \textbf{0.2146} & 22.1167 & \textbf{0.7561} & 3.9944 & 0.0374 & 0.3507 & 0.2822 \\
        \rowcolor{lightgreen}
        \cellcolor{white} & \textbf{Ours} & 0.2519 & 0.2098 & 22.7080 & 0.7901 & \textbf{6.0557} & \textbf{0.0772} & \textbf{0.4293} & \textbf{0.3304} \\
        \midrule
        \multirow{5}{*}{CogVideoX-5B}
        & Baseline & \textbf{0.2627} & \textbf{0.2140} & \textbf{23.4035} & 0.8243 & 2.6547 & 0.0384 & 0.3109 & 0.2415 \\
        & CADS & 0.2534 & 0.2116 & 22.7322 & \textbf{0.7651} & 2.9304 & 0.0494 & \textbf{0.4470} & 0.2884 \\
        & IG & 0.2539 & 0.2105 & 22.6066 & 0.7665 & \textbf{2.9594} & 0.0512 & 0.4368 & 0.2955 \\
        & VideoGuide & 0.2693 & 0.2176 & 22.9130 & 0.8717 & 1.5847 & 0.0216 & 0.2692 & 0.1774 \\
        \rowcolor{lightgreen}
        \cellcolor{white} & \textbf{Ours} & 0.2550 & 0.2115 & 22.7820 & 0.7666 & 2.9415 & \textbf{0.0522} & 0.4360 & \textbf{0.2976} \\
        \bottomrule
    \end{tabular}%
    }
    \vspace{-0.33cm}
\end{table}

\begin{table}[t]
    \centering
    \caption{\textbf{Quantitative comparison on T2V-CompBench}. ``Baseline'' indicates negative-prompted CFG. We observe that MotionCFG significantly improves temporal consistency and motion dynamics compared to existing guidance techniques. Upward arrows ($\uparrow$) indicate higher is better, while downward arrows ($\downarrow$) indicate lower is better.}
    \label{tab:t2vcomp_results}

    \resizebox{\textwidth}{!}{%
    \begin{tabular}{ll cccccccc}
        \toprule
        \multirow{2}{*}{\textbf{Model}} & \multirow{2}{*}{\textbf{Method}} & \multicolumn{2}{c}{\textbf{Framewise}} & \textbf{Full Video} & \multicolumn{5}{c}{\textbf{Dynamics / Temporal}} \\
        \cmidrule(lr){3-4} \cmidrule(lr){5-5} \cmidrule(lr){6-10}
        & & CLIPScore$\uparrow$ & PickScore$\uparrow$ & X-CLIP$\uparrow$ & SSIM$\downarrow$ & FLOW$\uparrow$ & Info DINO$\uparrow$ & DINO Segm.$\uparrow$ & DEVIL$\uparrow$ \\
        \midrule
        \multirow{6}{*}{Wan2.1-1.3B}
        & Baseline & 0.2651 & 0.2108 & 22.7719 & 0.7720 & 3.0671 & 0.0463 & 0.3517 & 0.2528 \\
        & CADS & \textbf{0.2654} & \textbf{0.2127} & \textbf{24.1668} & 0.7018 & 1.7996 & 0.0348 & 0.3536 & 0.2022 \\
        & IG & 0.2552 & 0.2090 & 23.9607 & 0.6785 & 4.1116 & 0.0601 & 0.4761 & 0.3077 \\
        & VideoGuide & 0.2675 & 0.2113 & 23.3390 & 0.6581  & 4.1287 & 0.0459 & 0.3992 & 0.2585  \\
        & SG & 0.2631 & 0.2088 & 22.4492 & 0.6581 & 4.1287 & 0.0459 & 0.3992 &  0.2579\\
        \rowcolor{lightgreen}
        \cellcolor{white} & \textbf{Ours} & 0.2550 & 0.2085 & 23.7080 & \textbf{0.6378} & \textbf{6.5136} & \textbf{0.0726} & \textbf{0.5034} & \textbf{0.3531} \\
        \midrule
        \multirow{4}{*}{Wan2.1-14B}
        & Baseline & \textbf{0.2755} & \textbf{0.2147} & 23.1646 & 0.8770 & 1.7147 & 0.0378 & 0.2561 & 0.1995 \\
        & CADS & 0.2630 & 0.2121 & 23.1093  & 0.7853 & 3.5325 & 0.0438 & 0.3464 & 0.2286 \\
        & IG & 0.2579 & 0.2092 & \textbf{23.2270} & 0.8223 & 1.7266 & 0.0571 & 0.3893 & 0.2581 \\
        \rowcolor{lightgreen}
        \cellcolor{white} & \textbf{Ours} & 0.2593 & 0.2097 & 23.0819 & \textbf{0.8025} & \textbf{4.4839} & \textbf{0.0727} & \textbf{0.4213} & \textbf{0.3057} \\
        \bottomrule
    \end{tabular}%
    }
\vspace{-0.33cm}
\end{table}

\subsection{Main Results}
\label{sec:main_results}

\noindent \textbf{Qualitative Evaluation.}
Figure~\ref{fig:qualitative} compares generated videos of MotionCFG against the baselines. Standard CFG with negative prompting (``Baseline'') often produces nearly static frames, while CADS and Interval Guidance introduce artifacts such as over-saturation or subject morphing. In contrast, MotionCFG generates vigorous motion trajectories, such as the structural collapse of the building and the galloping motion of the horse, while preserving subject consistency and visual fidelity. Since static frames cannot fully convey these temporal improvements, we provide high-resolution video comparisons in the \href{https://motion-cfg.vercel.app/}{project page}.

\definecolor{mcfg_red}{HTML}{E63946}
\definecolor{ig_orange}{HTML}{FFB800}
\definecolor{scfg_green}{HTML}{2A9D8F}
\definecolor{cads_blue}{HTML}{1D3557}

\definecolor{mcfg_red}{HTML}{E63946}
\definecolor{ig_orange}{HTML}{FFB800}
\definecolor{scfg_green}{HTML}{2A9D8F}
\definecolor{cads_blue}{HTML}{1D3557}

\noindent \textbf{Quantitative Results.} Tables \ref{tab:gemini_results} and \ref{tab:t2vcomp_results} present quantitative comparisons across all base models and prompt sets. MotionCFG consistently achieves high scores in FLOW, DINO Segm., and DEVIL, confirming its effectiveness in enhancing temporal dynamics at multiple granularities. Notably, on Wan2.1-14B, MotionCFG more than doubles the FLOW score of the baseline (6.06 vs.\ 2.95) while maintaining comparable X-CLIP. CADS tends to suppress motion (lower FLOW than baseline in most settings), and Interval Guidance offers moderate gains but at the cost of reduced CLIPScore. In contrast, MotionCFG amplifies motion dynamics with only marginal trade-offs in framewise fidelity metrics, demonstrating a favorable balance between dynamics and visual quality.

\begin{figure}[t]
    \centering
    \includegraphics[width=1.0\linewidth]{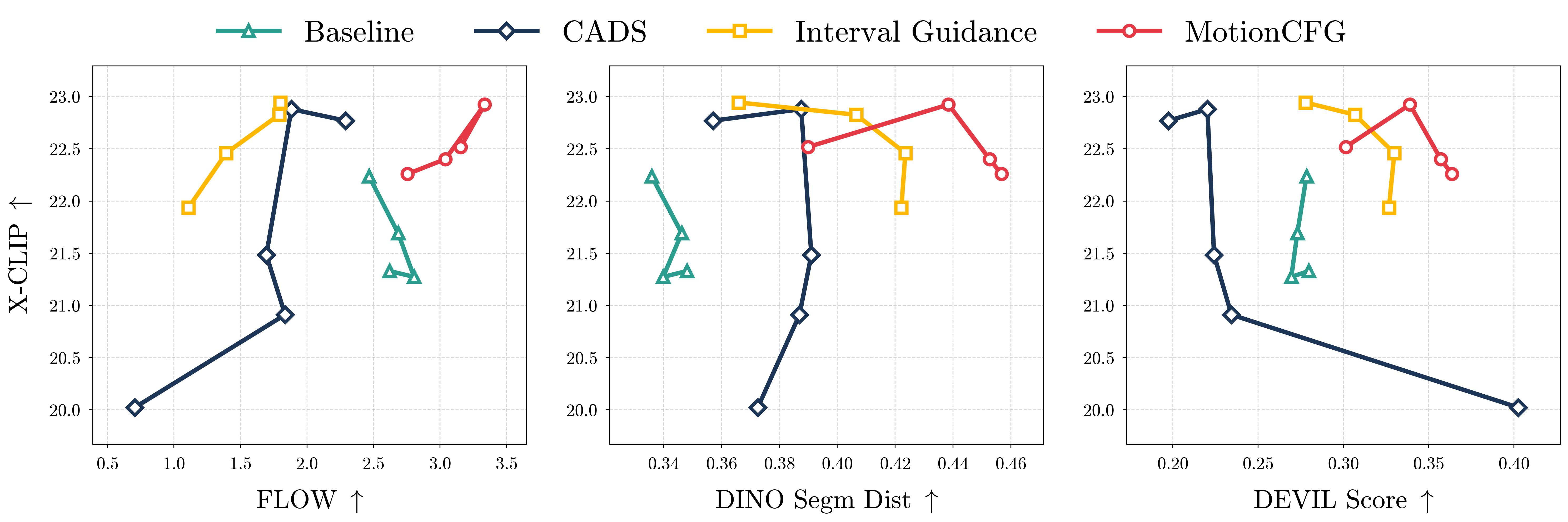}

    \caption{\textbf{Trade-off analysis between text fidelity and motion dynamics.} Each point corresponds to a different hyperparameter choice (\eg, ratio $\tau$). MotionCFG (\textcolor{mcfg_red}{red}) achieves higher motion scores (FLOW, Dino Segm Dist, DEVIL) with minimal X-CLIP degradation, while baselines either sacrifice fidelity or fail to improve dynamics.}

    \label{fig:tradeoff}
\end{figure}

Figure~\ref{fig:tradeoff} visualizes the text fidelity--motion dynamics trade-off by sweeping each method's hyperparameters. Observe that MotionCFG occupies the upper-right region across all three dynamics metrics, achieving stronger motion with minimal X-CLIP degradation. Baseline and Interval Guidance cluster in a narrow range, indicating limited controllability over dynamics. CADS can reach high FLOW and DINO Segm Dist values, but only at a severe X-CLIP cost; its curve drops steeply as dynamics increase. In contrast, MotionCFG's Pareto frontier remains nearly flat, demonstrating that motion intensity can be scaled without sacrificing semantic fidelity.

\subsection{Ablation Studies}
\label{sec:ablation}

\begin{wraptable}[13]{r}{0.55\textwidth}
\vspace{-0.2cm}
\centering
\caption{\textbf{Ablation on perturbation target.} 
We compare null/negative prompting with perturbations applied to text embeddings at non-motion or motion-related token positions. Perturbing motion-related embeddings yields the strongest dynamics while preserving competitive text alignment.}
\label{tab:ablation_perturbation}
\footnotesize
\setlength{\tabcolsep}{3pt}
\resizebox{0.55\textwidth}{!}{%
\begin{tabular}{lcccc}
\toprule
\textbf{Method} & \textbf{X-CLIP} $\uparrow$ & \textbf{I-DINO} $\uparrow$ & \textbf{D-Segm.} $\uparrow$ & \textbf{DEVIL} $\uparrow$ \\
\midrule
Null-Prompted & \textbf{22.2696} & 0.0597 & 0.3594 & 0.3083 \\
Neg-Prompted & 22.2128 & 0.0555 & 0.3842 & 0.3069 \\
Non-motion ($\mathcal{I}_{\text{motion}}^c$) & 21.9169 & \underline{0.0805} & \underline{0.4208} & \underline{0.3320} \\
Motion ($\mathcal{I}_{\text{motion}}$) & \underline{22.2222} & \textbf{0.0887} & \textbf{0.4990} & \textbf{0.4001} \\
\bottomrule
\end{tabular}%
}
\vspace{-1em}
\end{wraptable}

Table~\ref{tab:ablation_perturbation} isolates the effect of the perturbation target in the text embedding space. Null- and negative-prompted baselines show limited dynamics, indicating that replacing the unconditional branch with a generic static negative prompt is insufficient to resolve motion ambiguity. Perturbing embeddings at non-motion token positions improves several dynamic metrics, but causes a noticeable drop in X-CLIP, suggesting that perturbations outside the motion subspace disrupt static semantics such as object identity or scene layout. In contrast, perturbing embeddings at motion-related token positions consistently improves dynamic metrics while maintaining competitive text alignment, yielding the strongest fidelity--dynamics trade-off. Additional ablations including LLM-refined positive/negative prompts, $\sigma_c$, $\tau$, and other factors are provided in Section~\ref{sec:apdx_ablations}.

\subsection{Extension to Object Counting}
\label{sec:counting}

\begin{figure}[t]
    \centering
    \textbf{Prompt:} \textit{``\textcolor{red}{Three} horses gallop across the beach.''} \\[0.2cm]
    \vspace{-0.05cm}
    
    \begin{subfigure}[b]{0.48\textwidth}
        \centering
        \includegraphics[width=\textwidth]{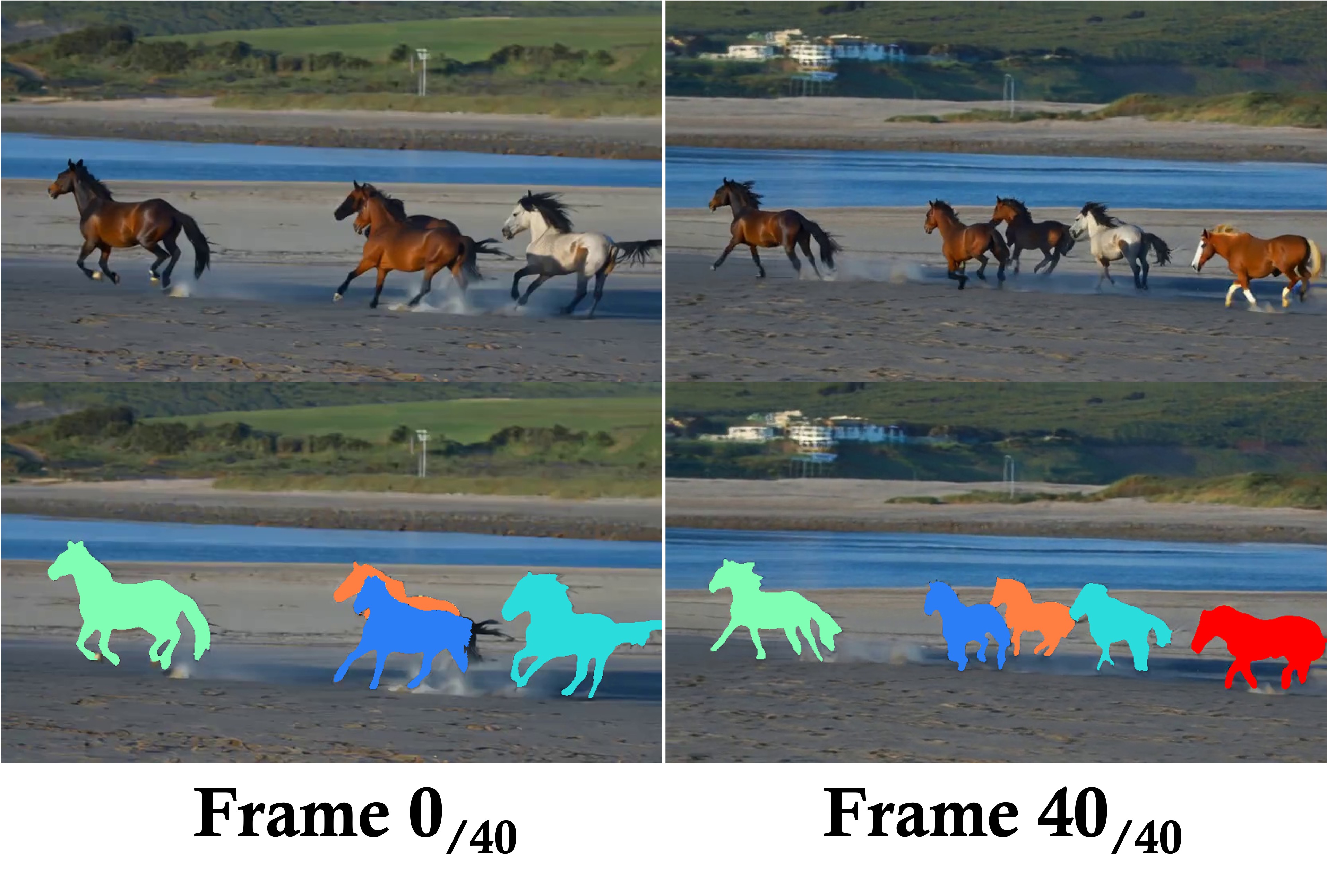}
        \vspace{-0.5cm}
        \caption{Interval Guidance}
        \label{fig:horse_ig}
    \end{subfigure}
    \hfill
    \begin{subfigure}[b]{0.48\textwidth}
        \centering
        \includegraphics[width=\textwidth]{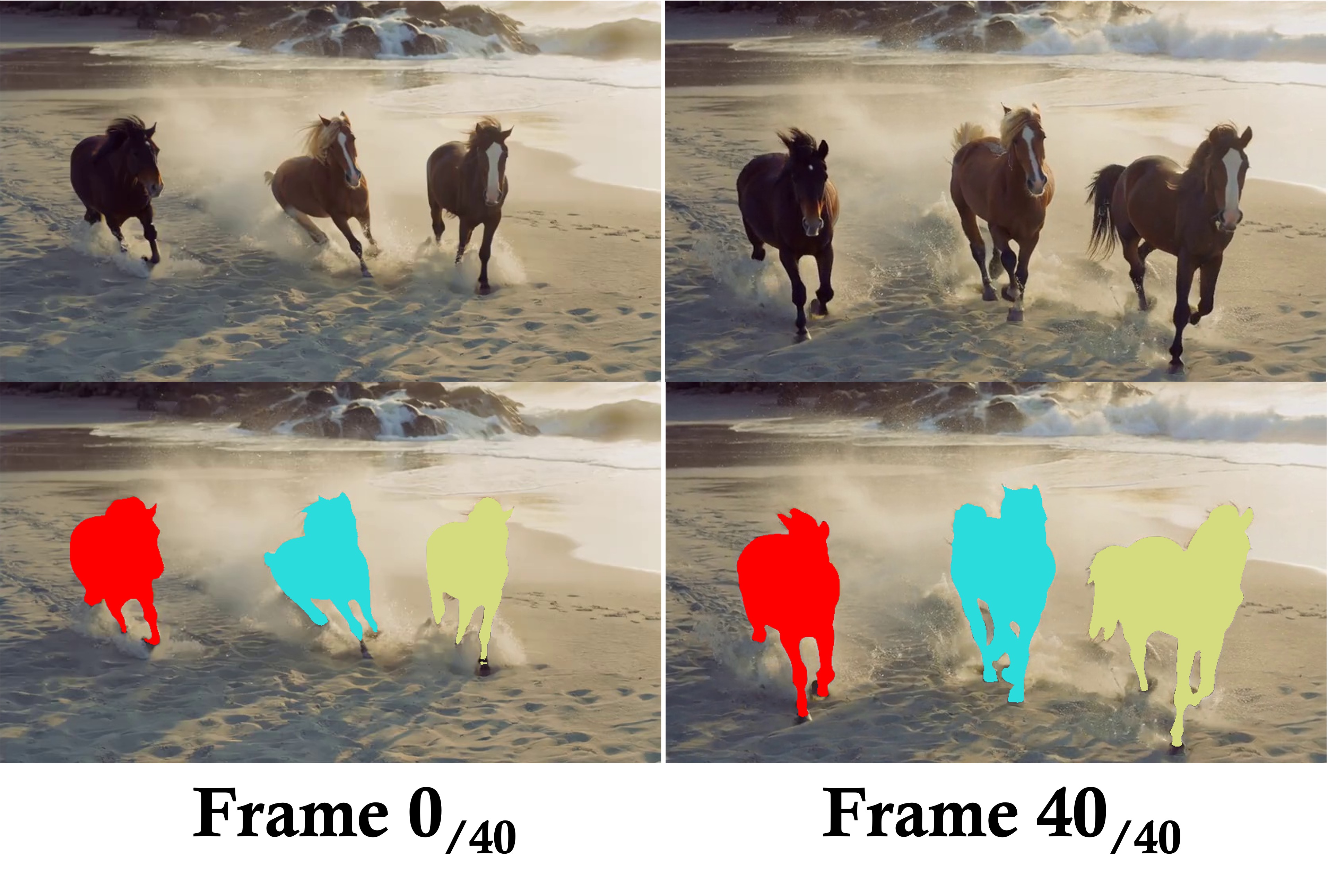}
        \vspace{-0.5cm}
        \caption{Ours}
        \label{fig:horse_ours}
    \end{subfigure}
    
    \caption{\textbf{Qualitative comparison of subject count consistency.} While the Interval Guidance (a) hallucinated an additional horse, our method (b) consistently maintained the requested count of three horses throughout the video.}
    \label{fig:horse_comparison}
    \vspace{-0.35cm}
\end{figure}

\begin{wraptable}{r}{0.48\linewidth}
    \vspace{-0.5cm}
    \centering
    \caption{\textbf{Object counting performance.} MotionCFG configurations show lower MAE and RMSE.}
    \label{tab:counting}
    \resizebox{\linewidth}{!}{
    \begin{tabular}{lcc}
        \toprule
        \textbf{Method} & \textbf{MAE} $\downarrow$ & \textbf{RMSE} $\downarrow$ \\
        \midrule
        Baseline ($w=6.0$) & 6.2543 & 8.3558 \\
        IG ($w=6.0, r=0.1$) & 4.8911 & 12.0896 \\
        \midrule
        Ours ($\tau=0.2, \sigma_c=0.1$) & 4.1411 & 11.5924 \\
        Ours ($\tau=0.4, \sigma_c=0.1$) & 4.2056 & 9.1235 \\
        Ours ($\tau=0.2, \sigma_c=0.5$) & \textbf{2.3952} & \textbf{3.6290} \\
        \bottomrule
    \end{tabular}
    }
\end{wraptable}

To demonstrate the versatility of our framework, we investigate whether the semantic perturbation mechanism of MotionCFG can be generalized beyond motion dynamics to other specific attributes, such as numerical quantity. In this experiment, we apply perturbations to the embeddings of quantity-related words instead of motion-related ones. We evaluate the object counting accuracy using the open-world counting framework proposed in CountVid \cite{AminiNaieni25}, calculating the Mean Absolute Error (MAE) and Root Mean Square Error (RMSE) between the generated videos and ground truth counts.

As summarized in Table~\ref{tab:counting}, MotionCFG demonstrates a clear advantage in numerical alignment compared to the baseline and Interval Guidance (IG). Even with preliminary hyperparameter settings, our method effectively reduces the counting error. For instance, the configuration with $w=6.0, \tau=0.2, \sigma_c=0.5$ achieves an MAE of 2.3952, which is a significant improvement over the baseline MAE of 6.2543. These results suggest that perturbing attribute-specific embeddings during the guidance process can effectively steer the generative model toward precise attribute formation. The variance in performance across different noise scales indicates a need for attribute-specific scheduling. Since different semantic properties likely emerge at different stages of the diffusion process, future work should explore optimal temporal windows and noise intensities tailored to specific attributes beyond motion and quantity.

\section{Conclusion}
\label{sec:conclusion}
We introduced MotionCFG, a training-free guidance mechanism that addresses motion ambiguity in text-to-video (T2V) generation. By constructing motion-perturbed conditions and steering generation away from them, MotionCFG selectively sharpens motion semantics while leaving other attributes intact. We further showed that this procedure admits an interpretable theoretical viewpoint as Semantic Laplacian Sharpening, a second-order correction restricted to the motion subspace of the condition. Combined with a piecewise guidance schedule, MotionCFG consistently improves motion dynamics across state-of-the-art T2V frameworks with negligible overhead, and the same principle generalizes to other ambiguous semantic attributes such as precise object numerosity.

\noindent\textbf{Limitations and Future Work.}
MotionCFG relies on an external LLM for automatic motion token identification, introducing a lightweight but non-negligible dependency; replacing it with attention-based saliency or similar attribution methods could further streamline the pipeline. Extending the framework to additional semantic axes and developing adaptive scheduling that automatically identifies the optimal intervention window are promising directions for future work.

Furthermore, our theoretical connection to Semantic Laplacian Sharpening holds in expectation, whereas practical guidance relies on a single stochastic perturbation at each denoising step. This approximation may introduce variability across samples, particularly for complex motion descriptions. Further investigation into variance reduction and more robust semantic conditioning could improve the consistency of motion generation.

\noindent\textbf{Broader Impact.}
By improving motion controllability in text-to-video generation, MotionCFG can benefit content creation and visual simulation. However, as with any advance in generative video quality, there is a risk of misuse for producing deceptive synthetic media. We encourage responsible deployment following the safety guidelines of the underlying generative models.

\section*{Acknowledgments}
This work was supported by the National Research Foundation of Korea under Grant RS-2024-00336454, the Advanced GPU Utilization Support Program funded by the Government of the Republic of Korea (Ministry of Science and ICT) (02-26-01-0404), the AI Computing Infrastructure Enhancement (GPU Rental Support) User Support Program funded by the Ministry of Science and ICT (MSIT) Republic of Korea (RQT-25-120217), and Artificial intelligence industrial convergence cluster development project funded by the Ministry of Science and ICT(MSIT, Korea) \& Gwangju Metropolitan City. This work was also supported by the National Research Foundation of Korea (NRF) grants funded by the Korean government (MSIT) (No. RS-2025-25463760; No. RS-2026-25541956, contribution rate: 15\%). 

%
%
\bibliographystyle{plain}
\bibliography{main}

\newpage
\appendix

\section{Implementation Details}
\label{sec:imple_details}

\noindent \textbf{Hyperparameter Settings.} For MotionCFG, we set the motion guidance scale to $\omega = 6.0$, the scheduling ratio to $\tau \in [0.1, 0.2]$, and the perturbation magnitude to $\sigma_c \in [0.1, 0.5]$. All approaches, including standard CFG, CADS, and Interval Guidance, share the same base guidance scale $\omega_{\text{std}} = 6.0$ for fair comparison.

\begin{promptbox}{Instruction Template for Motion Token Identification}
messages = [
    {"role": "system", "content": 
        "You are an AI that identifies ACTION VERBS in a sentence.\n"
        "Given indexed words like [i]word, output ONLY the indices of VERBS that describe physical actions or movements.\n"
        "DO NOT select: nouns, articles (a, an, the), pronouns (it, its, he, she), prepositions (in, on, at), adjectives.\n"
        "If no action verbs exist, return [].\n"
        "Output format: JSON list of indices only, nothing else."
    },
    # Few-shot: Explicit actions
    {"role": "user", "content": "Text: [0]A [1]cat [2]jumps [3]over [4]the [5]fence"},
    {"role": "assistant", "content": "[2]"},
    
    # Few-shot: Multiple actions
    {"role": "user", "content": "Text: [0]Birds [1]fly [2]and [3]sing [4]in [5]the [6]morning"},
    {"role": "assistant", "content": "[1, 3]"},
    
    # Few-shot: No actions (nouns only)
    {"role": "user", "content": "Text: [0]A [1]red [2]ball [3]on [4]the [5]table"},
    {"role": "assistant", "content": "[]"},
    
    # Few-shot: Exclude articles/pronouns
    {"role": "user", "content": "Text: [0]The [1]dog [2]sleeps [3]under [4]a [5]tree"},
    {"role": "assistant", "content": "[2]"},

    {"role": "user", "content": f"Text: {indexed_text}"}
]
\end{promptbox}

\noindent
\begin{minipage}[t]{0.57\linewidth}
\noindent \textbf{Delayed Scheduling for CogVideoX.}
Unlike Wan2.1, CogVideoX exhibits severe artifacts when MotionCFG is applied from the very first denoising step. As shown in Figure~\ref{fig:cog_tweedie}, the Tweedie estimate at $t=T$ is dominated by dataset-specific biases (\eg, POND5 watermarks) rather than meaningful semantic structure. Injecting motion perturbation at this stage may amplify these artifacts instead of enhancing dynamics. To address this, we employ a delayed schedule: the first few steps use only standard conditional prediction ($\omega_{\text{std}}=1.0$) to establish a clean spatial layout, after which MotionCFG is applied over the subsequent interval. This simple modification successfully enhances motion dynamics while preventing early-stage artifact amplification.
\end{minipage}
\hfill
\begin{minipage}[t]{0.39\linewidth}
\vspace{-0.5em}
\centering
\includegraphics[width=\linewidth]{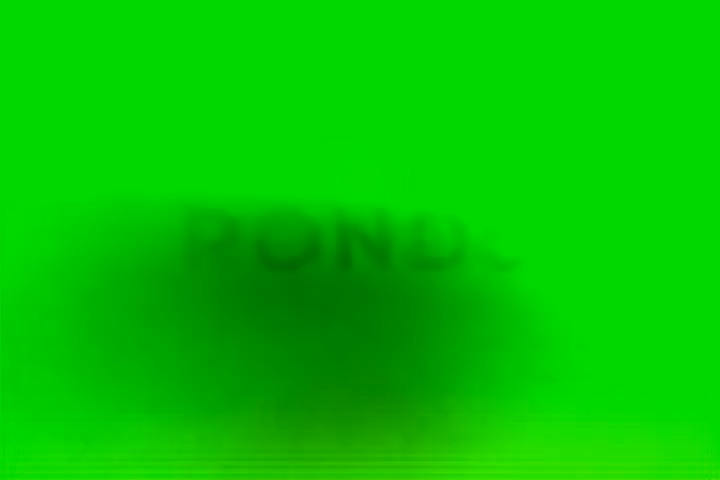}
\captionof{figure}{\textbf{CogVideoX Tweedie estimate at $t=T$.} The initial prediction is dominated by watermark-like artifacts.}
\label{fig:cog_tweedie}
\end{minipage}

\noindent \textbf{Motion Token Extraction Prompt.}

We use Qwen2.5-7B-Instruct to automate motion token identification. The input prompt is tokenized with positional indices (\eg, \texttt{[0]A [1]cat [2]jumps}), and the model is instructed to return indices of physical action verbs via few-shot in-context learning. Since modern text encoders (\eg, T5) often split a single word into multiple sub-word tokens, we locate the exact character span of each extracted verb and cross-reference it with the tokenizer's offset mapping. This yields the precise sub-word token indices, which form our active target set $\mathcal{I}_{motion}$ for the subsequent semantic perturbation. The full instruction template is shown in previous page.

\noindent \textbf{Standard Negative Prompts.}
To encourage high-quality generation and heavily penalize low-quality or motionless artifacts, we employ a comprehensive standard negative prompt across all baselines. The exact text of the negative prompt used in our evaluations is detailed below.

\begin{promptbox}{Standard Negative Prompt}
Bright tones, overexposed, static, blurred details, subtitles, style, works, paintings, images, static, overall gray, worst quality, low quality, JPEG compression residue, ugly, incomplete, extra fingers, poorly drawn hands, poorly drawn faces, deformed, disfigured, misshapen limbs, fused fingers, still picture, messy background, three legs, many people in the background, walking backwards
\end{promptbox}

\begin{table}[t]
\centering
\setlength{\tabcolsep}{18pt} 

\caption{\textbf{Examples from the Gemini-generated evaluation prompts.} Motion keywords are in \textbf{bold}.}

\label{tab:prompt_examples}
\resizebox{\textwidth}{!}{%
\begin{tabular}{@{}l p{10cm}@{}}
\toprule
\textbf{Category} & \textbf{Text Prompt} \\ \midrule
Animal & A wild horse \textbf{galloping} through a dusty canyon \\
Animal & A great white shark \textbf{swimming aggressively} in the deep ocean \\
Human & A ballet dancer \textbf{spinning} on stage under a spotlight \\
Human & A chef \textbf{chopping onions rapidly} on a wooden board \\
Vehicle & A red sports car \textbf{drifting} on a race track with smoke \\
Vehicle & A space rocket \textbf{launching} into the sky with fire \\
Nature & A volcano \textbf{erupting} with lava and ash \\
Physics & An old building \textbf{collapsing} into dust \\
Abstract & Colorful ink \textbf{spreading} in clear water \\ \bottomrule
\end{tabular}%
}
\end{table}

\noindent \textbf{Gemini-Generated Evaluation Prompts.}
Existing T2V benchmarks such as T2V-CompBench primarily evaluate spatial composition and attribute binding, often lacking prompts that demand vigorous physical motion. Models can thus achieve high alignment scores with near-static outputs. To address this gap, we curated a motion-intensive prompt set of 100 prompts generated via Gemini, spanning categories including animal behaviors, human actions, vehicle dynamics, and natural phenomena. Each prompt is centered around a distinct action verb describing high-magnitude motion. Representative examples are shown in Table~\ref{tab:prompt_examples}.

\section{Mathematical Derivation and Empirical Analysis}
\label{app:math_derivation}

\subsection{Proof of Proposition~\ref{prop:laplacian}}
\begin{manualproposition}{\ref{prop:laplacian}}
Let $\bn_t \sim \mathcal{N}(\mathbf{0}, \mathbf{I})$ be the noise realization in~\eqref{eq:perturbation}. Assuming $\mathbf{S}_\theta(\bz_t, \cdot)$ is three-times differentiable with respect to the condition, the expected guided score in~\eqref{eq:motioncfg_pw_score} satisfies
\begin{equation}
    \mathbb{E}_{\bn_t}\left[
    \tilde{\mathbf{S}}_\theta^*(\bz_t, \bc, \bc_{\text{pert},t})
    \right] \;=\; \mathbf{S}_\theta(\bz_t, \bc) \;-\; \frac{\omega\,\sigma_{\bc}^2}{2}\,\nabla_{\bc_m}^2\,\mathbf{S}_\theta(\bz_t, \bc) \;+\; O(\sigma_{\bc}^3),
\end{equation}
where $\nabla_{\bc_m}^2$ is the Laplacian with respect to $\bc_m$.
\end{manualproposition}
\begin{proof}
We first introduce a block decomposition of the text condition. Let $\bc \in \mathbb{R}^{L \times d}$ denote the text embedding, where $L$ is the number of tokens and $d$ is the embedding dimension. Given the motion-token index set $\mathcal{I}_{\text{motion}}$, we partition the condition into a motion block and a static block:
\begin{equation}
    \bc = (\bc_s, \bc_m), \qquad \bc_s \in \mathbb{R}^{d_s}, \quad \bc_m \in \mathbb{R}^{d_m},
    \label{eq:proof_block_decomp}
\end{equation}
where $\bc_m$ collects the vectorized embedding entries corresponding to motion-related tokens and $\bc_s$ contains the remaining static or contextual components. Under this notation, the perturbed condition in~\eqref{eq:perturbation} is written as $\bc_{\text{pert},t} = (\bc_s, \bc_m + \sigma_{\bc} \bn_t)$, leaving the static block untouched.

Next, we expand the perturbed score $\mathbf{S}_\theta(\bz_t, \bc_{\text{pert},t})$ around $\bc$ along the motion subspace via a second-order Taylor expansion. For each score coordinate $k \in \{1, \dots, d_z\}$,
\begin{equation}
\begin{aligned}
    S_\theta^{(k)}(\bz_t, \bc_{\text{pert},t})
    = S_\theta^{(k)}(\bz_t, \bc)
    &+ \sigma_{\bc}\, \nabla_{\bc_m} S_\theta^{(k)}(\bz_t, \bc)^\top \bn_t \\
    &+ \frac{\sigma_{\bc}^2}{2}\, \bn_t^\top \nabla_{\bc_m}^2 S_\theta^{(k)}(\bz_t, \bc)\, \bn_t
    + O(\sigma_{\bc}^3),
\end{aligned}
\label{eq:proof_taylor}
\end{equation}
where $\nabla_{\bc_m} S_\theta^{(k)} \in \mathbb{R}^{d_m}$ and $\nabla_{\bc_m}^2 S_\theta^{(k)} \in \mathbb{R}^{d_m \times d_m}$ are the gradient and Hessian of $S_\theta^{(k)}$ with respect to the motion block. The smoothness assumption (three-times differentiability) ensures the validity of this expansion.

Substituting~\eqref{eq:proof_taylor} into the MotionCFG update~\eqref{eq:motioncfg_pw_score} yields
\begin{equation}
\begin{aligned}
    \tilde{S}_\theta^{*,(k)}(\bz_t, \bc, \bc_{\text{pert},t})
    = S_\theta^{(k)}(\bz_t, \bc)
    &- \omega \sigma_{\bc}\, \nabla_{\bc_m} S_\theta^{(k)}(\bz_t, \bc)^\top \bn_t \\
    &- \frac{\omega \sigma_{\bc}^2}{2}\, \bn_t^\top \nabla_{\bc_m}^2 S_\theta^{(k)}(\bz_t, \bc)\, \bn_t
    + O(\sigma_{\bc}^3).
\end{aligned}
\label{eq:proof_update}
\end{equation}

We now take expectation over $\bn_t \sim \mathcal{N}(\mathbf{0}, \mathbf{I}_{d_m})$. The first-order term vanishes since $\mathbb{E}[\bn_t] = \mathbf{0}$. For the second-order term, we apply Hutchinson's identity~\cite{hutchinson1989stochastic}: for any matrix $A \in \mathbb{R}^{d_m \times d_m}$,
\begin{equation}
    \mathbb{E}_{\bn_t \sim \mathcal{N}(\mathbf{0}, \mathbf{I}_{d_m})}\left[\bn_t^\top A \bn_t\right]
    = \mathrm{Tr}(A).
    \label{eq:proof_hutchinson}
\end{equation}
Applying~\eqref{eq:proof_hutchinson} with $A = \nabla_{\bc_m}^2 S_\theta^{(k)}(\bz_t, \bc)$ gives
\begin{equation}
\begin{aligned}
    \mathbb{E}_{\bn_t}\left[\bn_t^\top \nabla_{\bc_m}^2 S_\theta^{(k)}(\bz_t, \bc)\, \bn_t\right]
    &= \mathrm{Tr}\!\left(\nabla_{\bc_m}^2 S_\theta^{(k)}(\bz_t, \bc)\right) \\
    &= \sum_{i=1}^{d_m} \frac{\partial^2 S_\theta^{(k)}(\bz_t, \bc)}{\partial (\bc_m)_i^2}
    \;=:\; \Delta_{\bc_m} S_\theta^{(k)}(\bz_t, \bc),
\end{aligned}
\label{eq:proof_laplacian_def}
\end{equation}
where the last equality defines the motion-condition Laplacian, recovering the standard scalar Laplacian operator restricted to the motion subspace.

Combining the above and applying the result component-wise to all $d_z$ score coordinates,
\begin{equation}
    \mathbb{E}_{\bn_t}\!\left[\tilde{\mathbf{S}}_\theta^*(\bz_t, \bc, \bc_{\text{pert},t})\right]
    = \mathbf{S}_\theta(\bz_t, \bc)
    - \frac{\omega \sigma_{\bc}^2}{2}\, \Delta_{\bc_m} \mathbf{S}_\theta(\bz_t, \bc)
    + O(\sigma_{\bc}^3).
\end{equation}
Identifying $\nabla_{\bc_m}^2 \mathbf{S}_\theta = \Delta_{\bc_m} \mathbf{S}_\theta$ (the trace of the motion-block Hessian, as defined in~\eqref{eq:proof_laplacian_def}) completes the proof.
\end{proof}

\subsection{Empirical Alignment with the Motion-Coordinate Laplacian}
\label{app:laplacian_alignment}

We complement Proposition~\ref{prop:laplacian} with two empirical
analyses on Wan2.1-T2V-1.3B. First, we compare the MotionCFG correction
with an independently sampled symmetric finite-difference estimate
along the motion-token coordinates. Second, we examine the local
motion-coordinate curvature encountered along CFG and MotionCFG
sampling trajectories.

\paragraph{Setup and estimators.}
For a fixed latent, timestep, and condition
$(\mathbf{z}_t,t,\mathbf{c})$, let
$\boldsymbol{\epsilon}_\theta(\mathbf{z}_t,t,\mathbf{c})$
denote the model prediction used in this analysis.
We define the Monte Carlo MotionCFG correction, before multiplication
by the guidance weight $\omega$, as
\begin{equation}
\mathbf{g}_{\mathrm{MC}}(K) = \boldsymbol{\epsilon}_\theta (\mathbf{z}_t,t,\mathbf{c}) -\frac{1}{K}\sum_{k=1}^{K} \boldsymbol{\epsilon}_\theta (\mathbf{z}_t,t,\mathbf{c} +\sigma_{\mathbf{c}}\mathbf{v}_k).
\label{eq:empirical_mc_correction}
\end{equation}
Each direction $\mathbf{v}_k$ is standard Gaussian on the selected
motion-token coordinates and zero elsewhere. Thus, $K=1$ corresponds
to the single-sample correction used by MotionCFG.

Using an independent set of Gaussian directions
$\{\mathbf{u}_j\}_{j=1}^{J}$ with the same support, we compute
\begin{equation}
\mathbf{g}_{\mathrm{Lap}} =-\frac{1}{2J}\sum_{j=1}^{J}\Big[ \boldsymbol{\epsilon}_\theta (\mathbf{z}_t,t,\mathbf{c} +\sigma_{\mathbf{c}}\mathbf{u}_j) -2\boldsymbol{\epsilon}_\theta (\mathbf{z}_t,t,\mathbf{c}) +\boldsymbol{\epsilon}_\theta (\mathbf{z}_t,t,\mathbf{c} -\sigma_{\mathbf{c}}\mathbf{u}_j) \Big].
\label{eq:empirical_fd_laplacian}
\end{equation}
The symmetric difference cancels the first-order term in each probe
direction. Under the small-perturbation expansion used in
Proposition~\ref{prop:laplacian},
\begin{equation}
\mathbb{E}[\mathbf{g}_{\mathrm{Lap}}] =-\frac{\sigma_{\mathbf{c}}^2}{2} \Delta_{\mathbf{c}_m} \boldsymbol{\epsilon}_\theta (\mathbf{z}_t,t,\mathbf{c}) +O(\sigma_{\mathbf{c}}^3).
\label{eq:empirical_scaled_laplacian}
\end{equation}
Thus, $\mathbf{g}_{\mathrm{Lap}}$ estimates the scaled negative
motion-coordinate Laplacian appearing in the correction.

Gaussian symmetry gives
$\mathbb{E}[\mathbf{g}_{\mathrm{MC}}(K)]
=\mathbb{E}[\mathbf{g}_{\mathrm{Lap}}]$.
This equality holds at finite perturbation scale, while the
interpretation of their shared expectation as a scaled Laplacian
relies on the small-perturbation expansion.
Our comparison therefore evaluates agreement between two
independently sampled estimators of the expected correction.

\paragraph{Results.}
Table~\ref{tab:laplacian_alignment} reports cosine similarity,
relative error, and norm ratio between the two estimates.
The single-sample correction is positively aligned with
$\mathbf{g}_{\mathrm{Lap}}$, with a cosine similarity of $0.732$,
although its relative error remains $0.916$.
Agreement improves as $K$ increases: at $K=128$, cosine similarity
reaches $0.992$, relative error decreases to $0.122$, and the norm
ratio approaches unity at $1.004$.
These results demonstrate close agreement in both direction and
magnitude after Monte Carlo averaging, while distinguishing this
expected behavior from the variability of an individual correction.

\begin{table*}[t]
\centering
\renewcommand{\arraystretch}{1.20}
\setlength{\tabcolsep}{4pt}

\noindent
\begin{minipage}[t]{0.48\textwidth}
\vspace{0pt}
\caption{\textbf{Alignment with the motion-coordinate Laplacian.}
Agreement between the Monte Carlo MotionCFG correction
$\mathbf{g}_{\mathrm{MC}}(K)$ and the independently sampled
symmetric finite-difference estimate $\mathbf{g}_{\mathrm{Lap}}$.
A norm ratio of one indicates matching magnitudes.
}
\label{tab:laplacian_alignment}
\end{minipage}%
\hfill
\begin{minipage}[t]{0.48\textwidth}
\vspace{0pt}
\caption{\textbf{Motion-coordinate curvature along sampling.}
Comparison of motion-coordinate finite-difference estimates
along CFG and MotionCFG trajectories at matched denoising steps.
Both branches share the initial latent, text condition,
and probe directions.
}
\label{tab:laplacian_trajectory}
\end{minipage}

\par\vspace{0.3em}
\noindent
\begin{minipage}[t]{0.48\textwidth}
\vspace{0pt}
\centering
\footnotesize
\begin{tabular*}{\linewidth}{@{\extracolsep{\fill}}rccc@{}}
\toprule
\rule[-2ex]{0pt}{6ex}$\vcenter{\hbox{$\boldsymbol{K}$}}$
& $\vcenter{\hbox{\shortstack{\textbf{Cosine}\\\textbf{sim.} $\uparrow$}}}$
& $\vcenter{\hbox{\shortstack{\textbf{Relative}\\\textbf{error} $\downarrow$}}}$
& $\vcenter{\hbox{\shortstack{\textbf{Norm}\\\textbf{ratio}}}}$ \\
\midrule
1   & 0.732 & 0.916 & 1.361 \\
4   & 0.901 & 0.475 & 1.129 \\
8   & 0.948 & 0.342 & 1.096 \\
128 & \textbf{0.992} & \textbf{0.122} & \textbf{1.004} \\
\bottomrule
\end{tabular*}
\end{minipage}%
\hfill
\begin{minipage}[t]{0.48\textwidth}
\vspace{0pt}
\centering
\footnotesize
\begin{tabular*}{\linewidth}{@{\extracolsep{\fill}}ccc@{}}
\toprule
\rule[-2ex]{0pt}{6ex}\textbf{Step}
& {\boldmath$
   \frac{\|\mathbf{g}_{\mathrm{Lap}}^{\mathrm{M}}\|_2}
        {\|\mathbf{g}_{\mathrm{Lap}}^{\mathrm{CFG}}\|_2}$}
& {\boldmath$
   \cos\!\left(
   \mathbf{g}_{\mathrm{Lap}}^{\mathrm{M}},
   \mathbf{g}_{\mathrm{Lap}}^{\mathrm{CFG}}
   \right)$} \\
\midrule
0  & 1.000 & 1.000 \\
5  & \textbf{2.211} & 0.079 \\
10 & 2.034 & 0.014 \\
35 & 1.236 & 0.014 \\
\bottomrule
\end{tabular*}
\end{minipage}
\end{table*}

\subsection{Motion-Coordinate Curvature along Sampling Trajectories}
\label{app:laplacian_trajectory}

We next investigate whether MotionCFG leads sampling through states
with different local motion-coordinate curvature.
Starting from the same initial latent and text condition, we run
standard CFG and MotionCFG and evaluate
~\eqref{eq:empirical_fd_laplacian} at matched denoising steps.
We denote the resulting estimates by
$\mathbf{g}_{\mathrm{Lap}}^{\mathrm{CFG}}$ and
$\mathbf{g}_{\mathrm{Lap}}^{\mathrm{M}}$, respectively.
At each matched step, we use the same Gaussian probe directions
for both branches to control variation due to probe sampling.

As shown in Table~\ref{tab:laplacian_trajectory}, the estimates coincide
at initialization and differ substantially after the trajectories
diverge. At step $5$, the MotionCFG estimate has $2.211$ times the
norm of the CFG estimate, with a cosine similarity of $0.079$.
The difference remains at step $35$, beyond the MotionCFG guidance
window, where the norm ratio is $1.236$ and cosine similarity is
$0.014$.

These measurements show that MotionCFG changes the local
motion-coordinate curvature encountered along the sampling
trajectory. The model parameters remain fixed; the differences arise
from evaluating the conditional prediction at different latent
states. Together with the fixed-state analysis above, these results
provide empirical evidence for the correction's finite-difference
interpretation and its effect on the subsequent sampling trajectory.

\section{Complementary Analyses and Discussions}
\label{sec:add_analysis}

\subsection{Toy Experiment}
\label{sec:toy_experiment}

\begin{wrapfigure}{r}{0.5\textwidth}
\vspace{-0.5cm}
\centering
\includegraphics[width=\linewidth]{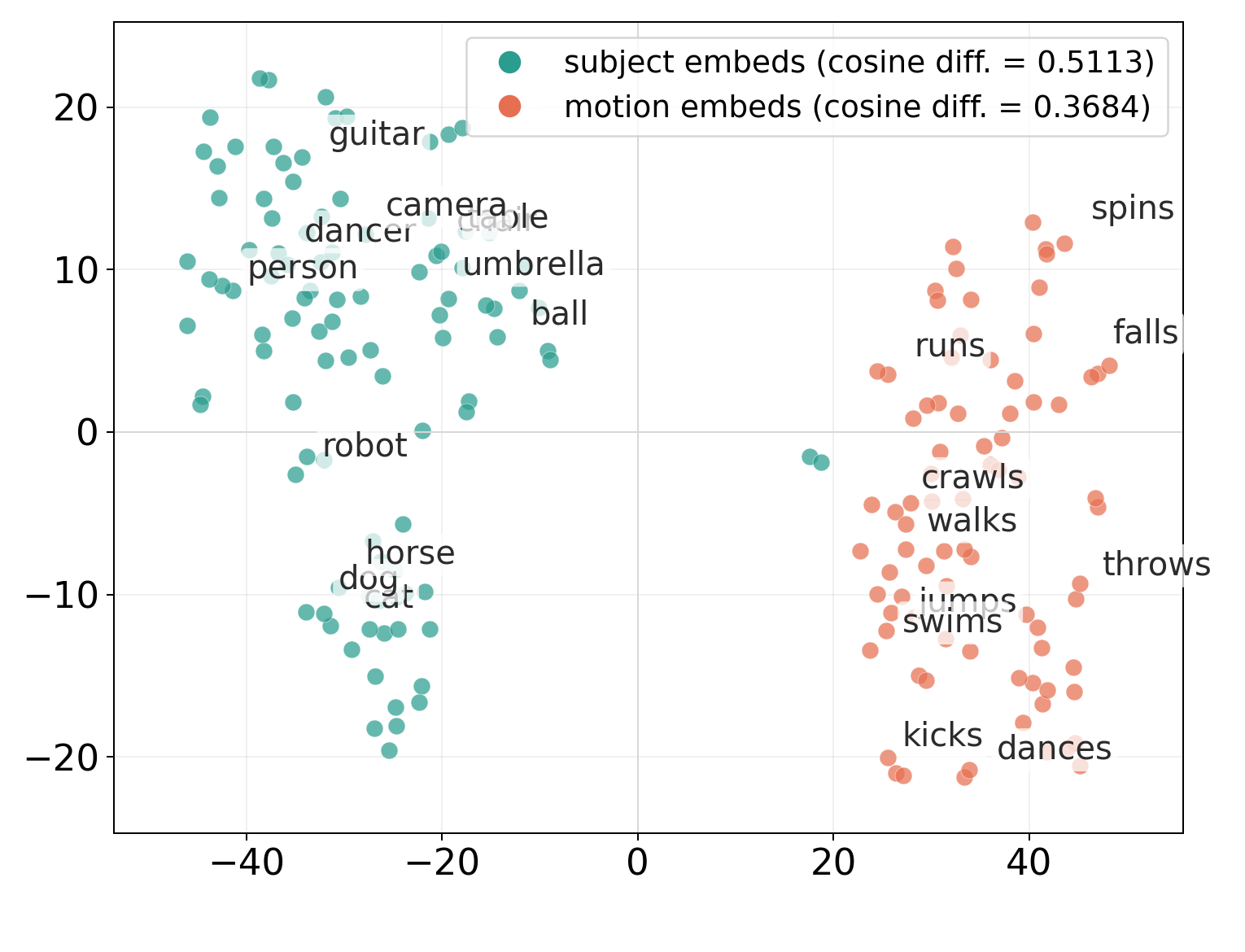}

\caption{
\textbf{T5 embedding geometry.} Subject tokens are more dispersed (avg.\ cosine distance 0.51) than motion tokens (0.37), confirming that motion semantics are more compact in the condition space.
}
\label{fig:embed_tsne}
\end{wrapfigure}

We conduct a two-dimensional flow-matching toy experiment to clarify how MotionCFG differs from standard CFG in the perspective of Semantic Laplacian Sharpening.

\noindent\textbf{Setup.}
We construct a conditional Gaussian mixture in $\mathbb{R}^2$, where $x_1$ is the motion axis and $x_2$ is the subject axis. A key design choice is the asymmetric mode separation: the motion modes are placed close together ($\mu_m = 0.8$) while the subject modes are placed farther apart ($\mu_s = 1.5$), with covariance $\boldsymbol{\Sigma}=\operatorname{diag}(1.0^2, 0.5^2)$. This reflects the structure of real T2V generation, where different motion interpretations (\eg, a strong jump vs.\ a subtle hop) are closer in feature space than different object identities (\eg, a cat vs.\ a dog), making the motion axis inherently harder to resolve. We further verify this asymmetry in the T5 encoder used by Wan2.1: as shown in Figure~\ref{fig:embed_tsne}, subject-related tokens exhibit larger average cosine distances than motion-related tokens, confirming that motion semantics occupy a more compact region in the text-embedding space. The resulting toy distribution has two motion modes that strongly overlap around $x_1=0$, creating a pronounced motion-ambiguous region, while the subject modes remain well separated.

For a condition $(c_m,c_s)$, $c_m$ controls the left/right motion preference and $c_s$ controls the lower/upper subject preference. 
The four mode scores are defined as $q_1=-c_m-c_s+2c_mc_s$, $q_2=-c_m+c_s-2c_mc_s$, $q_3=c_m-c_s-2c_mc_s$, and $q_4=c_m+c_s+2c_mc_s$. 
The mixture weights are obtained by a softmax over these scores, $\pi_j=\exp(q_j)/\sum_{k=1}^{4}\exp(q_k)$, and the conditional distribution is $p(\mathbf{x}\mid c_m,c_s)=\sum_{j=1}^{4}\pi_j\mathcal{N}(\mathbf{x};\boldsymbol{\mu}_j,\boldsymbol{\Sigma})$. 
In evaluation, we use a weak motion condition and a stronger subject condition, reflecting the case where subject identity is easier to resolve than motion.

We train a small conditional flow-matching model on this distribution. 
We sample $\mathbf{x}_0\sim\mathcal{N}(\mathbf{0},\mathbf{I})$ and $\mathbf{x}_1$ from the target mixture, train the velocity field along the straight path $\mathbf{x}_t=(1-t)\mathbf{x}_0+t\mathbf{x}_1$, and generate samples by Euler integration. 
At sampling time, standard CFG extrapolates away from the global null condition, while MotionCFG perturbs only the motion condition.

\begin{figure}[t]
\centering
\begin{subfigure}[t]{0.24\linewidth}
    \centering
    \includegraphics[width=\linewidth]{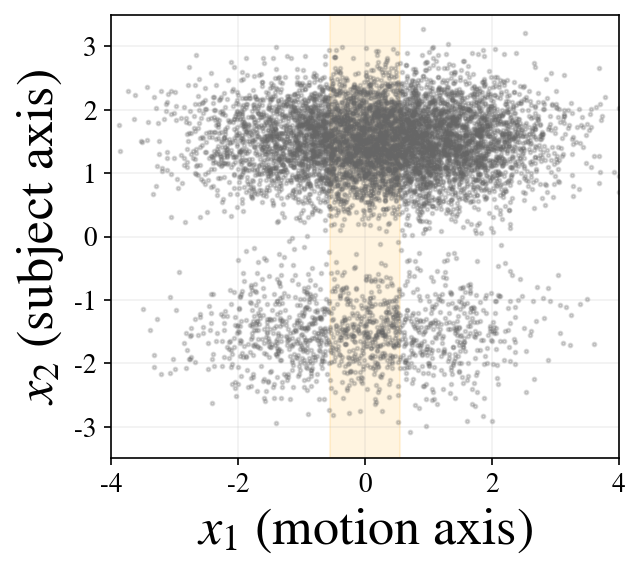}
    \caption{GT}
\end{subfigure}
\hfill
\begin{subfigure}[t]{0.24\linewidth}
    \centering
    \includegraphics[width=\linewidth]{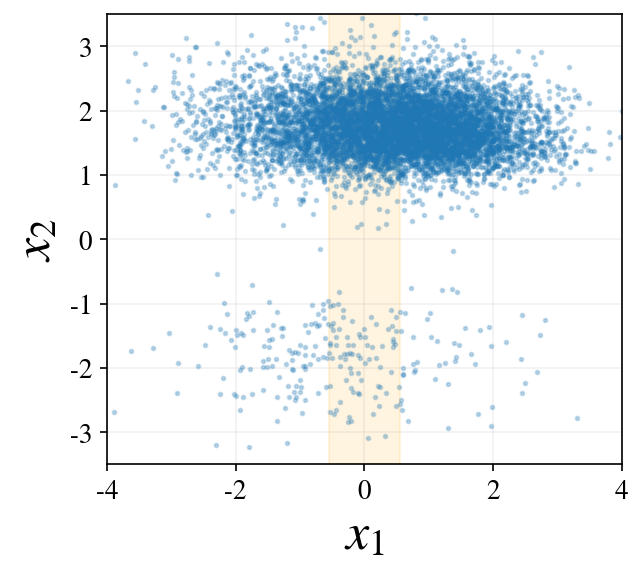}
    \caption{CFG}
\end{subfigure}
\hfill
\begin{subfigure}[t]{0.24\linewidth}
    \centering
    \includegraphics[width=\linewidth]{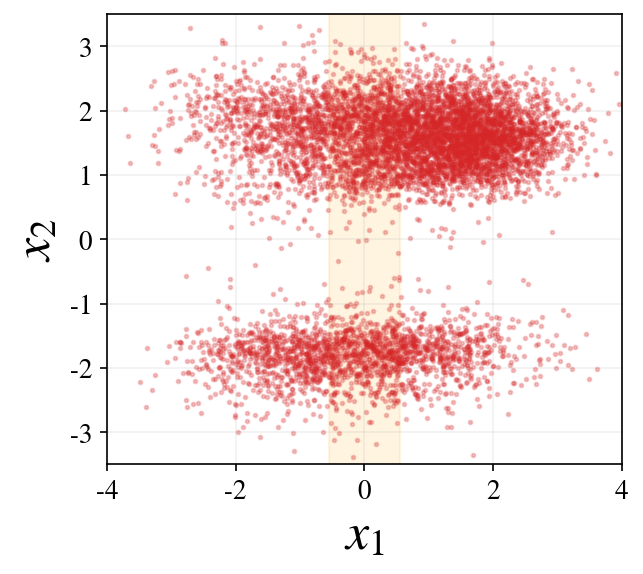}
    \caption{MotionCFG}
\end{subfigure}
\hfill
\begin{subfigure}[t]{0.24\linewidth}
    \centering
    \includegraphics[width=\linewidth]{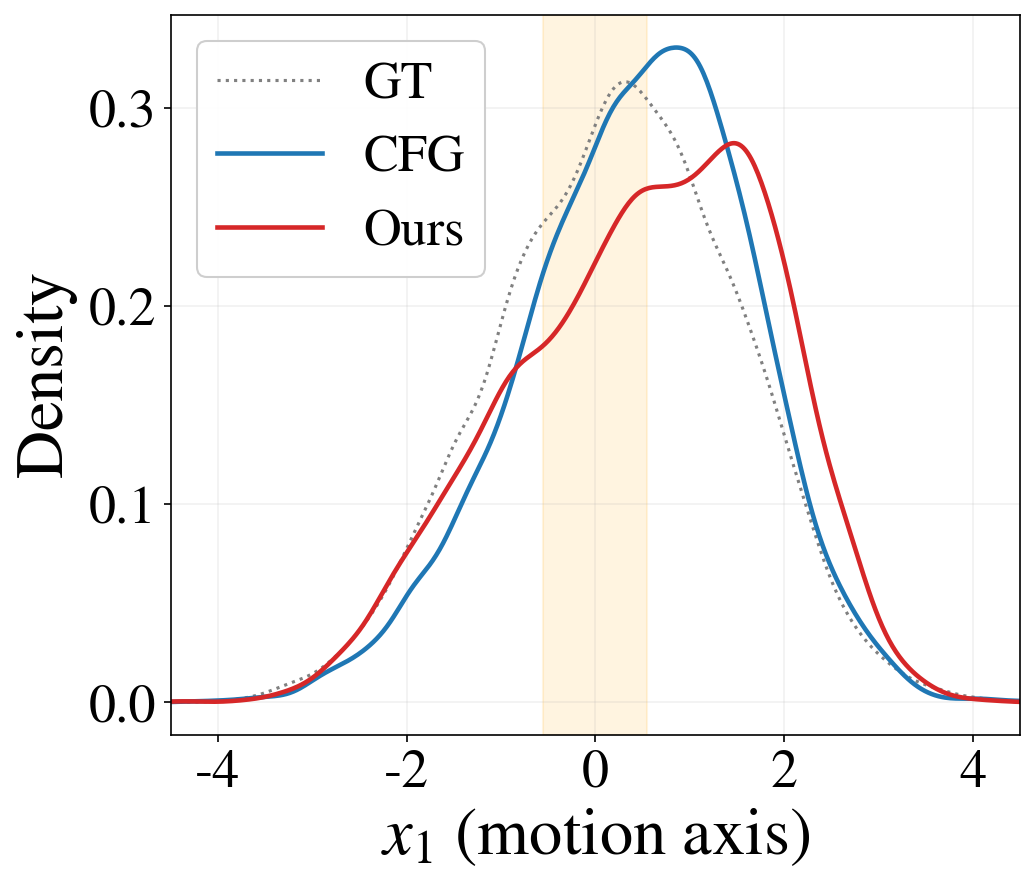}
    \caption{$x_1$ marginal}
\end{subfigure}
\caption{
\textbf{Flow-matching toy experiment.}
(a--c) 2D scatter plots of sampled distributions. The orange band marks the motion-ambiguous region around $x_1=0$. (b) Standard CFG  concentrates samples toward one dominant mode but fails to reduce mass in the ambiguous region.
(c) MotionCFG suppresses the ambiguous region while maintaining broader support along the motion axis. (d) Marginal density along the motion axis confirms that CFG (blue) leaves the ambiguous-region density comparable to GT (gray), while MotionCFG (red) clearly reduces it.
}
\label{fig:toy_motioncfg}
\end{figure}

\noindent\textbf{Qualitative behavior.}
Figure~\ref{fig:toy_motioncfg} visualizes the sampled distributions. 
The GT conditional distribution has substantial overlap along the motion axis, while the subject axis remains clearly separated. 
Standard CFG reduces some middle-region mass, but it does so by amplifying the entire condition against the global null branch. 
As a result, the samples become concentrated toward one dominant joint mode. 
This behavior is expected: CFG does not distinguish whether the amplified signal comes from motion or subject semantics, so a stronger subject signal can dominate the guidance direction.

MotionCFG behaves differently. 
Since the perturbed branch modifies only the motion condition, the resulting contrastive direction is localized to the under-resolved motion axis. 
The scatter plot shows that MotionCFG moves samples away from the ambiguous band around $x_1=0$ without collapsing as sharply as standard CFG. 
The histogram in Figure~\ref{fig:toy_motioncfg}(d) shows the same effect after marginalizing out the subject axis. 
The density inside the orange band corresponds to motion-ambiguous samples; MotionCFG lowers this density more clearly than standard CFG, while maintaining broader support along the motion axis.

\begin{wraptable}{r}{0.35\textwidth}
\vspace{-0.5cm}
\centering
\caption{
\textbf{Ambiguous-motion mass.}
$M_{\rm amb}$ is the fraction of samples inside the orange band around $x_1=0$. Lower is better.
}
\label{tab:toy_mamb}
\resizebox{0.8\linewidth}{!}{
\begin{tabular}{lc}
\toprule
\textbf{Method} & $\mathbf{M_{\rm amb}} \downarrow$ \\
\midrule
GT & 0.317 \\
Conditional & 0.322 \\
Standard CFG & 0.306 \\
\textbf{MotionCFG} & \textbf{0.233} \\
\bottomrule
\end{tabular}
}
\vspace{-2.0em}
\end{wraptable}

\noindent\textbf{Quantitative trend.}
Table~\ref{tab:toy_mamb} reports $M_{\rm amb}$, the fraction of samples in the motion-ambiguous band. 
Standard CFG only slightly reduces this value relative to the GT conditional distribution. 
MotionCFG provides a much stronger reduction, decreasing $M_{\rm amb}$ from $0.306$ to $0.233$ compared with standard CFG. 
Together with the scatter and histogram plots, this suggests that motion-local perturbation more directly suppresses uncertainty along the compact motion axis, whereas standard CFG performs a global sharpening that is more prone to single-mode concentration.

\begin{table}[t]
\centering

\begin{minipage}{0.48\linewidth}
\centering
\caption{\textbf{Effect of $\sigma_c$ (fixed $\tau=0.1$).} Increasing the scale provides scalable control over motion intensity.}
\label{tab:strength}
\resizebox{\linewidth}{!}{
\begin{tabular}{lccc}
\toprule
$\sigma_c$ & \textbf{X-CLIP} $\uparrow$ & \textbf{FLOW} $\uparrow$ & \textbf{DEVIL} $\uparrow$ \\
\midrule
0.01 & 23.0244 & 4.0177 & 0.3165 \\
0.03 & 22.8113 & 4.7649 & 0.3291 \\
0.05 & \textbf{23.3881} & 4.7320 & 0.3374 \\
0.10 & 22.6914 & \textbf{6.3951} & \textbf{0.3729} \\
\bottomrule
\end{tabular}
}

\end{minipage}
\hfill
\begin{minipage}{0.48\linewidth}
\centering
\caption{\textbf{Effect of Guidance Schedule ($\tau$).} Moderate intervals improve dynamics, while excessive $\tau$ degrades fidelity and motion.}
\label{tab:schedule}
\resizebox{\linewidth}{!}{
\begin{tabular}{lcccc}
\toprule
$\tau$ & \textbf{X-CLIP} $\uparrow$ & \textbf{FLOW} $\uparrow$ & \textbf{DEVIL} $\uparrow$ \\
\midrule
0.05 & \textbf{22.9430} & \textbf{1.8005} & 0.2779 \\
0.10 & 22.8258 & 1.7918  & 0.3069 \\
0.20 & 22.4572 & 1.3929  & \textbf{0.3300} \\
0.30 & 21.9369 & 1.1102 & 0.3268 \\
\bottomrule
\end{tabular}
}
\end{minipage}

\vspace{-0.5em}
\end{table}

\subsection{Additional Ablations}
\label{sec:apdx_ablations}

\noindent \textbf{Impact of Perturbation Strength ($\sigma_c$).}
We analyze the impact of the perturbation scale, denoted as $\sigma_c$, on the resulting motion dynamics. As summarized in Table~\ref{tab:strength}, increasing $\sigma_c$ significantly enhances the magnitude of generated motions, as indicated by the rise in FLOW and DEVIL Score. While higher $\sigma_c$ values introduce stronger perturbations to the motion-related embeddings, the X-CLIP score remains remarkably stable. This suggests that $\sigma_c$ serves as a reliable hyperparameter for scaling motion intensity while preserving the core semantic integrity.

\noindent \textbf{Effect of Guidance Schedule ($\tau$).}
We investigate the influence of the guidance schedule, controlled by the timestep ratio $\tau$, which determines the interval over which the semantic perturbation is applied. As shown in Table~\ref{tab:schedule}, increasing $\tau$ from $0.05$ to $0.20$ notably enhances high-level semantic dynamics, improving the DEVIL Score from $0.2779$ to $0.3300$ and DINO Segm. from $0.3660$ to $0.4236$. However, extending the perturbation interval too far ($\tau=0.30$) leads to a performance drop across most metrics. Notably, excessive $\tau$ causes a significant degradation in text alignment (X-CLIP drops to $21.9369$) and reduces raw pixel-level motion (FLOW decreases from $1.8005$ to $1.1102$). This indicates that while an appropriate application window is essential for developing structural motion, injecting perturbations over an excessively long schedule corrupts the spatial semantics and ultimately limits overall dynamic fidelity.

\noindent \textbf{Influence of LLM Size.}
A potential concern is MotionCFG's dependency on the accuracy of the LLM used for motion token extraction. To evaluate this, we ablate across the Qwen2.5 family (0.5B--7B) on our full evaluation set. As shown in Table~\ref{tab:llm_robustness}, smaller models occasionally misidentify tokens, \eg, Qwen2.5-0.5B with non-verbs such as ``a'' and ``nearby''. Despite such errors, MotionCFG still improves dynamics across all model sizes, with even the 0.5B variant outperforming the baseline (DEVIL: 0.319 vs.\ 0.279). Performance improves monotonically with LLM accuracy, and Qwen2.5-7B achieves the best overall trade-off, confirming that MotionCFG is robust to imprecise extraction and does not require a large-scale LLM to be effective.

\begin{table}[t]
\centering
\caption{\textbf{Robustness to LLM size for motion token extraction.}
The example column shows each model's extraction for the prompt: \textit{``An elephant sprays water with its trunk, a lion sitting nearby.''}}
\label{tab:llm_robustness}

\resizebox{1.0\linewidth}{!}{%
\begin{tabular}{l c c c c c}
\toprule
\textbf{Method} & \textbf{Extraction Example} & \textbf{X-CLIP}$\uparrow$ & \textbf{DINO Segm.}$\uparrow$ & \textbf{Info DINO}$\uparrow$ & \textbf{DEVIL}$\uparrow$ \\
\midrule
Standard CFG & - & 22.2369 & 0.3359 & 0.0568  & 0.2785 \\
\midrule
Qwen2.5-0.5B & ``sprays'', ``a'', ``nearby'' & 22.5443 & 0.4114 & 0.0751 & 0.3186 \\
Qwen2.5-3B & ``sprays'' & 22.8157 & 0.4342 & 0.0751 & 0.3334 \\
\rowcolor{lightgreen} 
\textbf{Qwen2.5-7B} & \textbf{``sprays''} & \textbf{22.9234}  & \textbf{0.4385} & \textbf{0.0765} & \textbf{0.3391} \\
\bottomrule
\end{tabular}%
}
\end{table}

\noindent \textbf{Comparison with LLM-based Prompt Engineering.}
We further compare MotionCFG with LLM-based prompt engineering on motion-related attribute prompts from T2V-CompBench. Specifically, we consider two prompt-level baselines: $\mathcal{P}_{\rm pos}$, where an LLM rewrites the original prompt into a more motion-explicit positive prompt, and $\mathcal{P}_{\rm neg}$, where an LLM generates an explicit negative prompt to discourage static or weak-motion outputs. As shown in Table~\ref{tab:prompt_engineering_t2vcomp}, prompt-level rewriting can improve certain motion metrics, but often sacrifices text-video alignment. In contrast, MotionCFG keeps the original prompt fixed and perturbs only motion-related text embeddings, yielding a more favorable fidelity--dynamics trade-off.

\noindent \textbf{Effect of Perturbation Locations.} 
We compare two intervention locations: the input token embeddings before the T5 Transformer and the final contextual embeddings supplied to the video denoiser. We evaluate $50$ prompts using paired initial seeds and identical settings ($\omega=6.0$, $\tau=0.2$, and $\sigma_{\mathbf{c}}=0.1$); only the perturbation location changes.

\begin{table}[t]
\centering
\caption{
\textbf{Comparison with LLM-based prompt engineering on motion-related T2V-CompBench prompts.}
Prompt-level rewriting can improve some motion metrics, but often degrades text-video alignment. MotionCFG achieves a stronger fidelity--dynamics trade-off by perturbing motion-related text embeddings directly.
}
\label{tab:prompt_engineering_t2vcomp}
\footnotesize
\setlength{\tabcolsep}{4pt}
\resizebox{0.95\linewidth}{!}{%
\begin{tabular}{lccccc}
\toprule
\textbf{Method} 
& \textbf{X-CLIP} $\uparrow$
& \textbf{FLOW} $\uparrow$
& \textbf{D-Segm.} $\uparrow$
& \textbf{I-DINO} $\uparrow$
& \textbf{DEVIL} $\uparrow$ \\
\midrule
LLM-refined $\mathcal{P}_{\rm pos}$ & 20.2686 & 1.9578 & 0.3664 & 0.0586 & 0.2547 \\
LLM-generated $\mathcal{P}_{\rm neg}$ & 17.4975 & \textbf{2.5841} & 0.4312 & \textbf{0.0686} & \textbf{0.2847} \\
MotionCFG (Qwen2.5-7B) & \underline{20.3305} & \underline{2.3076} & \underline{0.4802} & 0.0595 & 0.2690 \\
MotionCFG (Manual) & \textbf{21.1557} & 2.2613 & \textbf{0.4909} & \underline{0.0624} & \underline{0.2804} \\
\bottomrule
\end{tabular}%
}
\end{table}

\begin{table}[t]
\centering
\caption{\textbf{Ablation on perturbation location.}
We compare perturbations applied to initial T5 input embeddings and final contextual T5 embeddings using $50$ prompts with paired initial seeds and identical settings. 
}
\label{tab:perturbation_location}
\footnotesize
\setlength{\tabcolsep}{4pt}
\resizebox{0.95\linewidth}{!}{%
\begin{tabular}{lcccccc}
\toprule
\textbf{Perturbation location}
& \textbf{X-CLIP} $\uparrow$
& \textbf{SSIM} $\downarrow$
& \textbf{FLOW} $\uparrow$
& \textbf{D-Segm.} $\uparrow$
& \textbf{I-DINO} $\uparrow$
& \textbf{DEVIL} $\uparrow$ \\
\midrule
Initial T5 input
& \textbf{0.2354} & 0.8003 & 2.2231 & 0.4259 & 0.0566 & 0.2646 \\
Final T5 output
& 0.2314 & \textbf{0.7518} & \textbf{6.8930}
& \textbf{0.4570} & \textbf{0.0668} & \textbf{0.3192} \\
\bottomrule
\end{tabular}%
}
\end{table}

As shown in Table~\ref{tab:perturbation_location}, perturbing the final contextual embeddings yields higher FLOW, DINO Segm.\ Dist., Info-DINO, and DEVIL scores, together with lower SSIM. Perturbing the final embeddings acts directly on the contextual rows received by the denoiser and allows the original text representation to be cached and reused. In contrast, applying fresh perturbations to the initial input embeddings requires rerunning the text encoder at every active MotionCFG step. These results support the final contextual embeddings as the intervention location under the evaluated settings, with the additional computational benefit of avoiding repeated text-encoder evaluation.

\subsection{Computational Overhead}
\label{apdx:computation}

\begin{wraptable}{r}{0.4\textwidth}
    \vspace{-1cm}
    \centering
    \caption{\textbf{Complexity analysis.} Average end-to-end sampling time (sec/video). MotionCFG achieves superior temporal dynamics with negligible computational overhead.}
    \label{tab:complexity_tradeoff}
    \resizebox{\linewidth}{!}{%
    \begin{tabular}{lccc}
        \toprule
        \textbf{Method} & \textbf{Time}$\downarrow$ & \textbf{X-CLIP}$\uparrow$ & \textbf{DEVIL}$\uparrow$ \\
        \midrule
        Baseline & 76.339 & 22.2128 & 0.3069 \\
        CADS & 76.710 & 22.7687 & 0.1973 \\
        IG & \textbf{73.305} & 23.0677 & 0.3147 \\
        FlowMo & 106.490 & 22.4004 & 0.3032 \\
        \rowcolor{lightgreen}
        \textbf{Ours} & 77.310 & \textbf{23.3881} & \textbf{0.3374} \\
        \bottomrule
    \end{tabular}%
    }
    \vspace{-0.5cm}
\end{wraptable}

Table~\ref{tab:complexity_tradeoff} reports end-to-end sampling time per video on a single NVIDIA B200 GPU using Wan2.1-T2V-1.3B. 
Note that MotionCFG adds only $\sim$1 second over the baseline, as it requires no architectural modification, \ie, introducing only a lightweight perturbation of text embeddings during the early denoising steps. 
The LLM inference for motion token extraction is negligible relative to the video diffusion process. 

In contrast, FlowMo~\cite{flowmo2025} performs step-wise gradient-based optimization in the latent space, resulting in significantly higher inference time (e.g., $1.48\times$ slower than the baseline). Despite this minimal overhead, MotionCFG achieves the highest X-CLIP and DEVIL scores among all methods.

\subsection{Failure Cases}
\label{sec:failure_cases}

\begin{wrapfigure}{r}{0.45\textwidth}
    \vspace{-1.0cm}
    \centering
    \includegraphics[width=\linewidth]{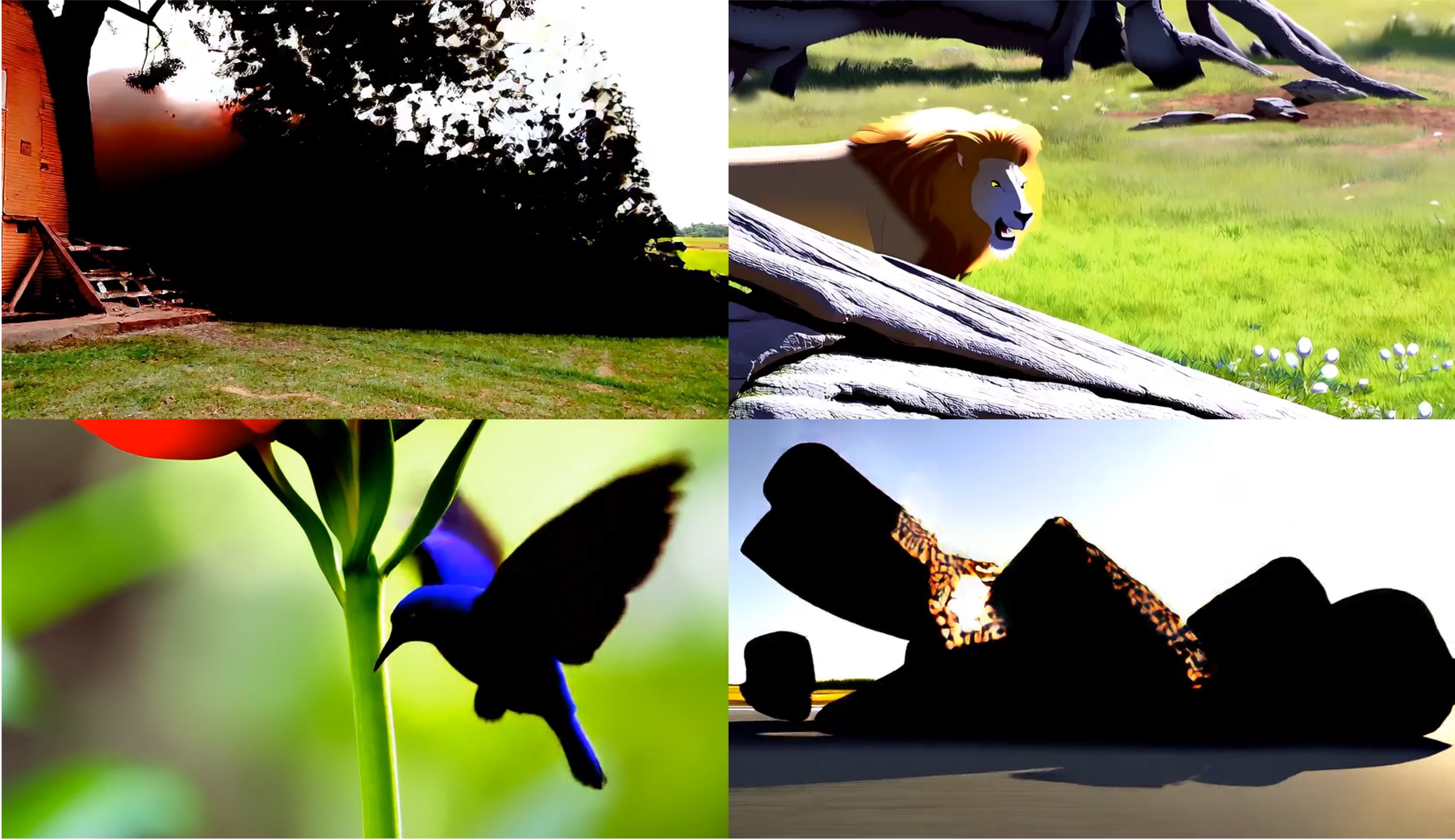}
    \vspace{-0.5cm}
    \caption{\textbf{Failure cases under extreme perturbation} ($\sigma_c = 3.0$). The perturbed condition falls far outside the valid semantic space, causing severe artifacts including over-saturation, loss of spatial detail, and black silhouetting.}
    \label{fig:failure_cases}
    \vspace{-1cm}
\end{wrapfigure}
While MotionCFG performs reliably under the demonstrated settings ($\sigma_c < 1.0$), extreme perturbation scales can cause generation failures. As shown in Figure~\ref{fig:failure_cases}, setting $\sigma_c = 3.0$ pushes $\bc_{\text{pert},t}$ far from the valid semantic space, and extrapolating away from this out-of-distribution anchor drives the latent trajectory into degenerate regions. This manifests as extreme contrast, loss of fine-grained detail, and large black silhouetted patches. These results confirm that $\sigma_c$ should be kept moderate (\eg, $\sigma_c < 1.0$) to ensure the perturbed condition remains a meaningful contrastive signal rather than uninformative noise.

\section{User Study}
\label{sec:user_study}

\begin{table*}[t]
\centering
\footnotesize
\caption{\textbf{User study results.}
Relative preference (\%) and absolute success rates (\%) of MotionCFG compared with baseline methods.
Preference scores are obtained from 32 participants evaluating visual quality and text alignment (C1) and task-specific performance (C2/C3).
Absolute correctness is evaluated by 36 additional participants on physical plausibility and action correctness.
The both-fail columns report cases where neither method succeeds.}
\label{tab:user_study}

\setlength{\tabcolsep}{5pt}
\begin{tabular}{llccrrrrrr}
\toprule
& & \multicolumn{2}{c}{\textbf{Preference (\%)}} &
\multicolumn{2}{c}{\textbf{Physical Success (\%)}} &
\multicolumn{2}{c}{\textbf{Action Success (\%)}} &
\multicolumn{2}{c}{\textbf{Both Fail (\%)}} \\
\cmidrule(lr){3-4}
\cmidrule(lr){5-6}
\cmidrule(lr){7-8}
\cmidrule(lr){9-10}
\textbf{Task} & \textbf{Comparison}
& C1 & C2/C3
& Other & Ours
& Other & Ours
& Physical & Action \\
\midrule
\multirow{3}{*}{\textit{Motion}}
& vs.\ Baseline
& \textbf{77.42} & \textbf{93.55}
& 31.11 & \textbf{82.23}
& 35.55 & \textbf{86.11}
& 13.33 & 9.44 \\
& vs.\ CADS
& \textbf{79.35} & \textbf{89.68}
& 22.22 & \textbf{67.78}
& 28.33 & \textbf{77.78}
& 20.00 & 10.00 \\
& vs.\ IG
& \textbf{85.16} & \textbf{81.94}
& 28.89 & \textbf{69.45}
& 34.45 & \textbf{73.89}
& 25.56 & 18.33 \\
\midrule
\textit{Counting}
& vs.\ Baseline
& \textbf{77.42} & \textbf{69.68}
& -- & -- & -- & -- & -- & -- \\
\bottomrule
\end{tabular}
\end{table*}

We conducted two complementary user studies to evaluate MotionCFG in terms of relative preference and absolute correctness. In the preference study, 32 participants evaluated 20 pairs of generated videos based on visual quality and text alignment (C1), and task-specific performance: motion dynamics (C2) or counting accuracy (C3). The presentation order was randomized to reduce ordering bias.

As shown in Table~\ref{tab:user_study}, MotionCFG is consistently preferred across all comparisons. For motion generation, MotionCFG achieves a 93.55\% preference rate over the baseline in motion dynamics, while also outperforming CADS and IG. For counting, MotionCFG is preferred over the baseline in both visual quality and counting accuracy.

To further evaluate absolute correctness beyond relative preference, we conducted an additional study with 36 participants. Each comparison was classified into four mutually exclusive outcomes: both methods fail, only the compared method succeeds, only MotionCFG succeeds, or both methods succeed. Physical plausibility and action correctness were assessed separately, with absolute success defined as the sum of exclusive success and joint success.

\newpage
\section*{NeurIPS Paper Checklist}

\begin{enumerate}

\item {\bf Claims}
    \item[] Question: Do the main claims made in the abstract and introduction accurately reflect the paper's contributions and scope?
    \item[] Answer: \answerYes{}.
    \item[] Justification: The abstract and introduction state that MotionCFG is a training-free, inference-time guidance method for improving motion dynamics in text-to-video generation. The theoretical claim (Proposition~\ref{prop:laplacian} in Section~\ref{sec:theory}) is supported by a formal proof (Section~\ref{app:math_derivation}) and empirical validation (Section~\ref{sec:toy_experiment}). The empirical claims are supported by quantitative comparisons across multiple models (Section~\ref{sec:main_results}), ablation studies (Section~\ref{sec:ablation}), and qualitative examples, while the extension to object counting (Section~\ref{sec:counting}) is presented as an additional demonstration rather than the primary claim.
    

\item {\bf Limitations}
    \item[] Question: Does the paper discuss the limitations of the work performed by the authors?
    \item[] Answer: \answerYes{}.
    \item[] Justification: Limitations and future work are discussed at the end of Section~\ref{sec:conclusion}. Additional failure cases and sensitivity analyses are provided in Section~\ref{sec:add_analysis}.

\item {\bf Theory assumptions and proofs}
    \item[] Question: For each theoretical result, does the paper provide the full set of assumptions and a complete (and correct) proof?
    \item[] Answer: \answerYes{}.
    \item[] Justification: Proposition~\ref{prop:laplacian} states its assumptions explicitly (three-times differentiability of the score function with respect to the condition). The complete proof is provided in Section~\ref{app:math_derivation} with detailed derivation steps. We also discuss the practical limitations of the asymptotic result, including single-sample Monte Carlo variance and higher-order residuals, and provide empirical validation on synthetic data (Section~\ref{sec:toy_experiment}).

    \item {\bf Experimental result reproducibility}
    \item[] Question: Does the paper fully disclose all the information needed to reproduce the main experimental results of the paper to the extent that it affects the main claims and/or conclusions of the paper (regardless of whether the code and data are provided or not)?
    \item[] Answer: \answerYes{}.
    \item[] Justification: The paper specifies all hyperparameters ($\omega$, $\sigma_c$, $\tau$), base models (Wan2.1-1.3B/14B, CogVideoX-5B), prompt sets, baselines, and evaluation metrics. Implementation details including the LLM-based motion token extraction prompt, negative prompt templates, and computational overhead analysis are provided in Section~\ref{sec:imple_details}.

\item {\bf Open access to data and code}
    \item[] Question: Does the paper provide open access to the data and code, with sufficient instructions to faithfully reproduce the main experimental results, as described in supplemental material?
    \item[] Answer: \answerNo{}.
    \item[] Justification: The code and data are not publicly released at submission time to preserve anonymity. However, the method is training-free and requires no additional learned parameters. The paper provides full algorithmic details (Algorithm~\ref{alg:motioncfg}), hyperparameters, prompt construction protocol, and evaluation metrics sufficient to reproduce the main results. We plan to release the code upon acceptance.

\item {\bf Experimental setting/details}
    \item[] Question: Does the paper specify all the training and test details (e.g., data splits, hyperparameters, how they were chosen, type of optimizer) necessary to understand the results?
    \item[] Answer: \answerYes{}.
    \item[] Justification: The paper specifies the evaluated text-to-video models, prompt sets, baselines, metrics, guidance scales, perturbation strengths, and schedule ratios in Sections~\ref{sec:experiments} and~\ref{sec:imple_details}. Since MotionCFG is training-free, no optimizer or training split is required for the proposed method.    

\item {\bf Experiment statistical significance}
    \item[] Question: Does the paper report error bars suitably and correctly defined or other appropriate information about the statistical significance of the experiments?
    \item[] Answer: \answerNo{}.
    \item[] Justification: The current version reports aggregate metrics over benchmark prompt sets but does not include confidence intervals, error bars, or statistical significance tests. We omit them due to the high computational cost of repeated large-scale video generation, and report multiple metrics, ablations, and model-scale comparisons as complementary evidence.

\item {\bf Experiments compute resources}
    \item[] Question: For each experiment, does the paper provide sufficient information on the computer resources (type of compute workers, memory, time of execution) needed to reproduce the experiments?
    \item[] Answer: \answerYes{}.
    \item[] Justification: Section~\ref{apdx:computation} reports average end-to-end sampling time per video and the GPU used for the complexity analysis. It also compares the computational overhead of MotionCFG against standard CFG and other inference-time baselines.
    
\item {\bf Code of ethics}
    \item[] Question: Does the research conducted in the paper conform, in every respect, with the NeurIPS Code of Ethics \url{https://neurips.cc/public/EthicsGuidelines}?
    \item[] Answer: \answerYes{}.
    \item[] Justification: The work is an inference-time method evaluated on existing text-to-video models and benchmark or curated prompt sets. We do not collect private data or release a new high-risk pretrained model.

\item {\bf Broader impacts}
    \item[] Question: Does the paper discuss both potential positive societal impacts and negative societal impacts of the work performed?
    \item[] Answer: \answerYes{}.
    \item[] Justification: In Section~\ref{sec:conclusion}, the paper discusses the positive impact of improving controllability and motion fidelity in text-to-video generation, and also acknowledges limitations including reliance on an external LLM and the need for adaptive scheduling across semantic axes.
    
\item {\bf Safeguards}
    \item[] Question: Does the paper describe safeguards that have been put in place for responsible release of data or models that have a high risk for misuse (e.g., pre-trained language models, image generators, or scraped datasets)?
    \item[] Answer: \answerNA{}.
    \item[] Justification: The paper does not introduce or release a new pretrained image/video generator, scraped dataset, or large-scale model. Experiments are conducted using existing models, and responsible access is governed by the usage policies and safeguards of those underlying model providers.

\item {\bf Licenses for existing assets}
    \item[] Question: Are the creators or original owners of assets (e.g., code, data, models), used in the paper, properly credited and are the license and terms of use explicitly mentioned and properly respected?
    \item[] Answer: \answerYes{}.
    \item[] Justification: We use publicly available pretrained models and benchmarks under their respective licenses and terms of use, including Apache 2.0-licensed Wan2.1 and CogVideoX code, and research-use licensed CogVideoX-5B weights, T2V-CompBench, CountVid, and Qwen2.5. We cite the original sources for all assets and do not redistribute proprietary datasets or model weights.

\item {\bf New assets}
    \item[] Question: Are new assets introduced in the paper well documented and is the documentation provided alongside the assets?
    \item[] Answer: \answerYes{}.
    \item[] Justification: The paper introduces a curated motion-intensive prompt set generated to evaluate high-magnitude motion dynamics. Section~\ref{sec:imple_details} documents the prompt construction protocol, categories, and representative examples; no private or human-subject data are included.

\item {\bf Crowdsourcing and research with human subjects}
    \item[] Question: For crowdsourcing experiments and research with human subjects, does the paper include the full text of instructions given to participants and screenshots, if applicable, as well as details about compensation (if any)? 
    \item[] Answer: \answerYes{}.
    \item[] Justification: We conducted a user study with 32 participants who evaluated 20 pairs of generated videos on visual quality and task-specific performance (Section~\ref{sec:user_study}). Participation was voluntary and no personally identifiable information was collected. The evaluation details and settings are described in Section~\ref{sec:user_study}.

\item {\bf Institutional review board (IRB) approvals or equivalent for research with human subjects}
    \item[] Question: Does the paper describe potential risks incurred by study participants, whether such risks were disclosed to the subjects, and whether Institutional Review Board (IRB) approvals (or an equivalent approval/review based on the requirements of your country or institution) were obtained?
    \item[] Answer: \answerNA{}.
    \item[] Justification: The user study involved only preference judgments on generated videos, with no collection of personally identifiable information and no risk to participants. This type of evaluation is exempt from IRB review at our institution.

\item {\bf Declaration of LLM usage}
    \item[] Question: Does the paper describe the usage of LLMs if it is an important, original, or non-standard component of the core methods in this research? Note that if the LLM is used only for writing, editing, or formatting purposes and does \emph{not} impact the core methodology, scientific rigor, or originality of the research, declaration is not required.
    \item[] Answer: \answerYes{}.
    \item[] Justification: The paper explicitly describes the use of a lightweight LLM for motion token identification in the proposed pipeline. The supplementary material further evaluates robustness to different LLM sizes and extraction quality.

\end{enumerate}

\end{document}